\documentclass[10pt,twocolumn,letterpaper]{article}
\usepackage[pagenumbers]{cvpr}

\usepackage{microtype}

\newcommand{\retargettablefont}{\rmfamily}
\DeclareCaptionFont{retargettimes}{\retargettablefont}
\usepackage{multirow,pifont}
\DeclareUnicodeCharacter{03BB}{\ensuremath{\lambda}}
\DeclareUnicodeCharacter{0394}{\ensuremath{\Delta}}
\DeclareUnicodeCharacter{2192}{\textrightarrow}
\newenvironment{nolinenumbers}{}{}

\definecolor{cvprblue}{rgb}{0.21,0.49,0.74}
\usepackage[pagebackref,breaklinks,colorlinks,allcolors=cvprblue]{hyperref}
\graphicspath{{figures/}}

\def\confName{CVPR}
\def\confYear{2027}
\title{What Should EEG Be Aligned To? Brain-Grounded Visual Targets for EEG-to-Image Retrieval}
\author{Jiahui Chai \qquad Zhenyao Cui \qquad Dongrui Wu\thanks{Corresponding author: \texttt{drwu09@gmail.com}.}\\
Huazhong University of Science and Technology\\
{\tt\small jiahuichai@hust.edu.cn\qquad zycui@hust.edu.cn\qquad drwu09@gmail.com}}

\hypersetup{pdftitle={What Should EEG Be Aligned To? Brain-Grounded Visual Targets for EEG-to-Image Retrieval},pdfauthor={Jiahui Chai, Zhenyao Cui, Dongrui Wu}}
\begin{document}
\maketitle

\begin{abstract}
Decoding images from EEG provides a window into human visual perception. Existing approaches commonly align EEG with features from pretrained vision models. However, these models are optimized for machine-oriented objectives, and their representations may not be optimal targets for EEG decoding. A comparison across eight vision models shows that decodability declines toward the final layers, where the machine-oriented objective acts most directly.  Moreover, handcrafted descriptors, including Gabor filters, HOG, and color histograms, surprisingly achieve 76.4\% top-1 accuracy in 200-way within-subject retrieval, showing that simple visual properties can support strong EEG decoding. Building on these observations, we propose ReTarget, a dual-branch framework that combines explicit visual priors with EEG-guided visual adaptation. The explicit branch aligns EEG with handcrafted descriptors, while the adaptation branch uses fixed EEG representations as anchors to adapt a deep vision encoder. On THINGS-EEG, ReTarget achieves 94.77\% within-subject top-1 accuracy. Further analyses show that the gains from combining the two branches arise from complementary visual targets rather than more models, highlighting the importance of selecting and adapting visual representations for EEG decoding.
\end{abstract}

\section{Introduction}

\begin{figure}[t]
    \centering
    \includegraphics[width=\columnwidth,trim={96bp 95bp 242bp 16bp},clip]{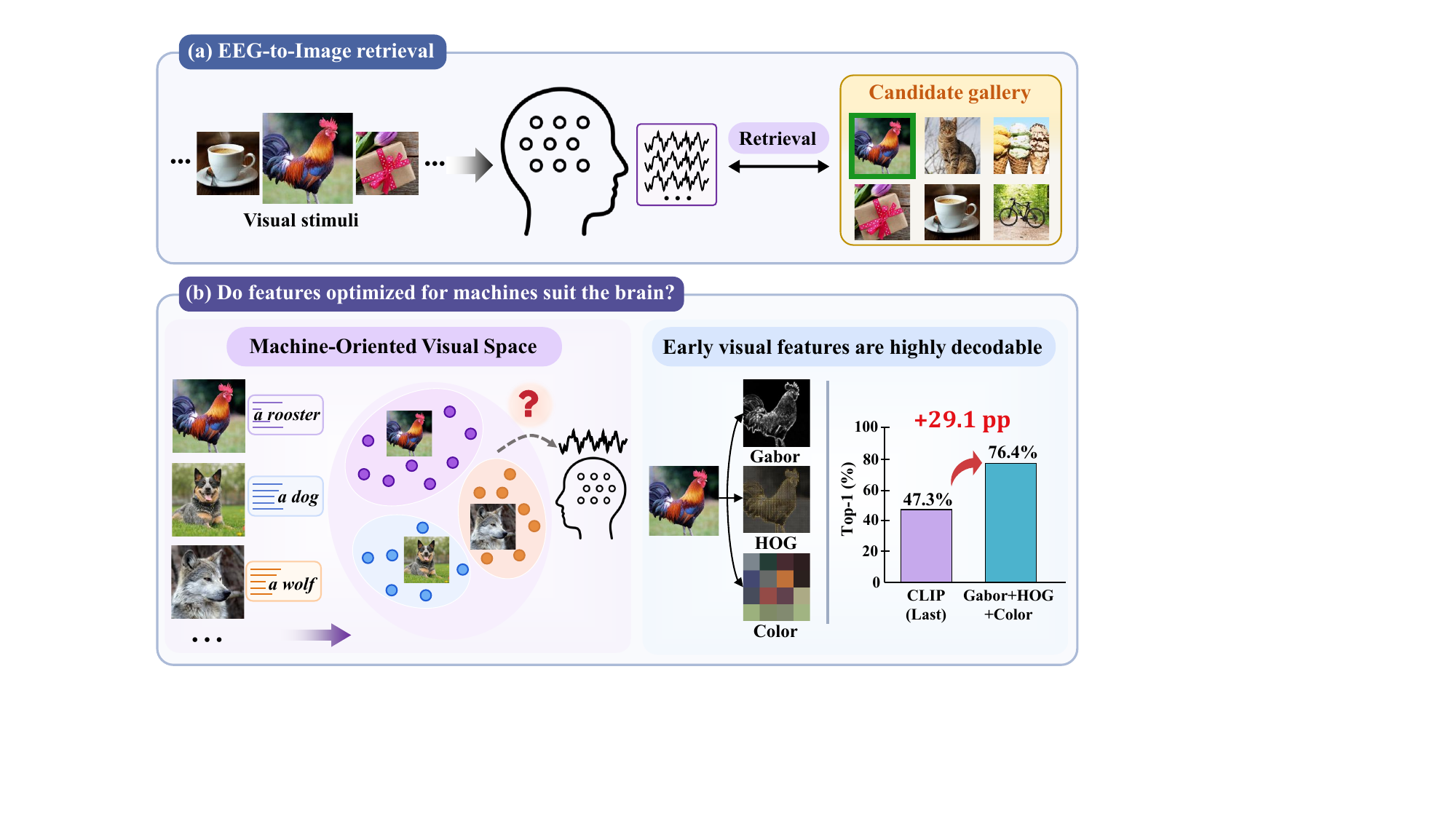}
    \caption{\textbf{What should EEG be aligned to?}
    (a) EEG-to-image retrieval identifies the viewed image (green box) in a candidate gallery.
    (b) Common visual targets come from encoders trained for machine tasks. Aligning EEG to descriptors of low-level visual properties instead yields higher 200-way within-subject top-1 accuracy than the final-layer CLIP target (Figs.~\ref{fig:descriptor_ablation} and~\ref{fig:layerwise}).}
    \label{fig:intro}
\end{figure}

Decoding visual content from brain activity offers a way to study how the brain represents what we see~\cite{kay2008identifying,horikawa2013neural} and may support brain--computer interfaces for communication and rehabilitation~\cite{chaudhary2016brain}.
EEG-to-image retrieval pursues this goal non-invasively: given the EEG response to a viewed image, it identifies that image in a candidate gallery (Fig.~\ref{fig:intro}a).
Most existing EEG-based visual decoding methods~\cite{song2024nice,li2024atm,li2025neuralmcrl,wu2025ubp,zhang2026neurobridge} address this task with contrastive learning, which aligns EEG embeddings with features extracted by pretrained image encoders such as CLIP~\cite{radford2021clip}.
In this framework, the contrastive objective raises the similarity of positive EEG--image pairs and lowers that of negative pairs, encouraging the EEG encoder to separate images along the differences encoded by the visual features.
The visual target thus not only provides a reference for alignment but also shapes the learning direction of the EEG encoder through the image differences it emphasizes, and in turn affects the final decoding performance.

However, the visual targets adopted by existing methods mostly come from models pretrained for machine vision tasks, such as image classification~\cite{he2016resnet}, image--text contrastive learning~\cite{radford2021clip,cherti2023openclip}, and visual self-supervised learning~\cite{caron2021dino,he2022mae,tian2024synclr}.
Although these models perform well on the tasks they were trained for, their pretraining does not consider which visual information can be reliably decoded from EEG.
Representations optimized for machine tasks are therefore not necessarily the most suitable targets for EEG alignment.
This prompts us to revisit a more fundamental question: what visual representation should EEG be aligned to?

To explore this question, we systematically compare the decoding performance of different visual targets within a unified EEG--image alignment framework.
Interestingly, a target built from handcrafted low-level descriptors such as Gabor filters~\cite{manjunath1996gabor}, HOG~\cite{dalal2005hog}, and color histograms~\cite{swain1991color} reaches 76.40\% top-1 accuracy in 200-way within-subject retrieval without relying on any deep visual pretraining (Fig.~\ref{fig:intro}b).
This result shows that simple low-level descriptors can provide a stable and effective alignment target for EEG decoding.
Consistent with Shallow Alignment~\cite{du2026shallow}, we further observe that intermediate layers outperform the final layer in each of the eight pretrained vision encoders we evaluate.
This indicates that even within a pretrained deep model, the final-layer representation is not necessarily the optimal target for EEG decoding.
Together, these observations suggest that improving EEG decoding need not focus only on fitting EEG to existing visual representations.
The visual target itself can also be redesigned to better capture what EEG reliably reflects.

Following this direction, we propose ReTarget (Fig.~\ref{fig:framework}), a dual-branch framework that combines explicit visual priors with EEG-anchored visual adaptation.
The explicit visual prior branch directly builds a fixed visual target from Gabor, HOG, and color descriptors, allowing the EEG encoder to align with these highly decodable visual properties.
For deep visual targets, selecting a suitable intermediate layer can improve decoding~\cite{du2026shallow,tang2026hcf}, but this strategy remains confined to the fixed representations that the pretrained model already provides.
We therefore go a step further and use EEG to guide the adaptation of the vision encoder itself.
Specifically, the EEG-anchored visual adaptation branch first learns the correspondence between EEG and deep visual features.
It then fixes the learned EEG representation as an anchor and fine-tunes the vision encoder with LoRA~\cite{hu2022lora}, so that the visual representation actively aligns with EEG.
At test time, we rank gallery images by a weighted sum of the similarity scores from the two branches.

Our main contributions are summarized as follows:
\begin{itemize}
    \item We systematically revisit visual targets for EEG decoding and show that representations optimized for machine tasks are not necessarily the most suitable alignment targets for EEG, as handcrafted low-level descriptors alone support strong retrieval performance.
    \item We propose ReTarget, a dual-branch framework combining explicit visual priors with EEG-anchored visual adaptation. Beyond selecting among frozen pretrained visual representations, the adaptation branch uses a learned EEG representation as a fixed anchor to adapt the vision encoder itself.
    \item ReTarget achieves 94.77\% within-subject and 30.00\% leave-one-subject-out top-1 accuracy on THINGS-EEG~\cite{gifford2022thingseeg2}. Controlled experiments show that its fusion gain stems mainly from the complementarity of explicit and deep visual targets rather than from the number of models.
\end{itemize}

\section{Related Work}

\paragraph{EEG-based visual decoding and retrieval.}
Contrastive alignment between EEG and pretrained image features underlies most recent EEG-to-image retrieval methods.
NICE~\cite{song2024nice} uses this alignment for zero-shot retrieval of unseen concepts, and ATM~\cite{li2024atm} designs a tailored EEG encoder for both retrieval and reconstruction.
Because the image features shape what the EEG encoder learns, recent work has gradually shifted toward the image side.
NeuroBridge~\cite{zhang2026neurobridge} augments images with blur, noise, low resolution, and mosaic before feature extraction.
Rajabi et al.~\cite{rajabi2025humanaligned} replace standard image encoders with versions fine-tuned on human similarity judgments~\cite{fu2023dreamsim,sundaram2024perceptual}.
Shallow Alignment~\cite{du2026shallow} and HCF~\cite{tang2026hcf} move the target to intermediate layers, and HCF further fuses several layers through learnable linear weights.
These methods transform the input image, replace the vision encoder, or change its readout layer, yet the vision encoder itself typically stays frozen and is not adjusted according to EEG.
Instead of selecting layers or learning projections on top of such features, we use the EEG representation as an anchor and adapt the visual encoder toward it.

\paragraph{Visual representations and perceptual alignment.}
Pretrained vision encoders learn visual representations through different training objectives.
CLIP~\cite{radford2021clip} and OpenCLIP~\cite{cherti2023openclip} learn image--text correspondence through contrastive pretraining on large-scale image--caption pairs.
In contrast, DINO~\cite{caron2021dino} learns visual features through self-distillation without labels, while MAE~\cite{he2022mae} reconstructs masked image patches.
SynCLR~\cite{tian2024synclr} performs contrastive learning on synthetic images, treating images generated from the same caption as positive pairs.
DreamSim~\cite{fu2023dreamsim} and subsequent work~\cite{sundaram2024perceptual} fine-tune pretrained vision encoders using human perceptual similarity judgments to obtain human-aligned visual representations.
Sec.~\ref{sec:layerwise} compares these encoders as EEG alignment targets.

\paragraph{Handcrafted visual models.}
Long before deep networks, handcrafted image models were used to characterize early visual responses and to recognize objects.
In early vision, two-dimensional Gabor functions give a reasonably accurate description of the spatial receptive fields of simple cells in cat striate cortex~\cite{jones1987gabor}.
Building on this receptive-field model, Kay et al.~\cite{kay2008identifying} modeled fMRI voxels in early visual areas with Gabor wavelets and used these models to identify which novel natural image a subject had seen.
For object recognition, HOG~\cite{dalal2005hog} represents local shape with contrast-normalized histograms of gradient orientations and was developed for human detection.
Color histograms~\cite{swain1991color} summarize the color distribution of an object and remain stable under occlusion and changes in view, making them an efficient cue for object indexing.
Unlike deep network features, each of these descriptors encodes known visual attributes, which makes it possible to ask directly which visual attributes EEG can decode.

\section{Revisiting Visual Targets for EEG}

\subsection{Preliminaries}
\label{sec:preliminaries}
All comparisons in this section share one retrieval framework, so that accuracy differences within each comparison can be attributed to the visual target.
Given paired EEG responses and images $\{(E_i, I_i)\}_{i=1}^{N}$, we describe each image $I_i$ with a visual target $g$ and encode each EEG response $E_i$ with an EEG encoder $f_E$.
Modality-specific projection heads $P_E$ and $P_I$ then map the resulting features into a shared embedding space:
\[
q_i = P_E\big(f_E(E_i)\big), \qquad v_i = P_I\big(g(I_i)\big).
\]
Following common practice in EEG-to-image retrieval~\cite{song2024nice},
we train with a symmetric InfoNCE loss~\cite{oord2018cpc,zhang2026neurobridge} that pulls matched EEG--image pairs together and pushes mismatched pairs apart within each mini-batch.
Let $s_{ij}=\mathrm{sim}(q_i,v_j)/\tau$, where $\tau$ is a fixed temperature.
\begin{equation}
\begin{aligned}
\mathcal{L}_{\mathrm{NCE}} &= -\frac{1}{2B}\sum_{i=1}^{B}
\Bigg[ \log\frac{\exp(s_{ii})}{\sum_{j=1}^{B}\exp(s_{ij})} \\
&\qquad\qquad\quad + \log\frac{\exp(s_{ii})}{\sum_{j=1}^{B}\exp(s_{ji})}
\Bigg],
\end{aligned}
\label{eq:align}
\end{equation}
where $B$ is the batch size.
At test time, an EEG query ranks all images in the candidate gallery $\mathcal{G}$ by the cosine similarity between $q$ and each $v$.
We fix a target, train $f_E$, $P_E$, and $P_I$ with Eq.~(\ref{eq:align}), and refer to the resulting top-1 retrieval accuracy as the decodability of that target.

\subsection{Explicit Visual Descriptors}
Most existing methods align EEG with features from large pretrained vision models, implicitly assuming that strong decoding requires deep, learned representations.
We question this assumption and ask how far EEG decoding can go when the target contains no learned representation at all.
To answer this, we replace the deep target with Gabor filters~\cite{manjunath1996gabor}, HOG~\cite{dalal2005hog}, and regional color histograms~\cite{swain1991color}, together with all their combinations.
For a controlled comparison, we fix the total dimensionality at 768 and train a separate EEG encoder for each target.
Despite their simplicity, these descriptors prove to be strong targets (Fig.~\ref{fig:descriptor_ablation}).
Gabor is the most decodable single descriptor.
Color is the weakest alone, yet it complements Gabor far more than HOG does.
Combining all three descriptors reaches 76.40\% top-1 accuracy without any visual pretraining, showing that visual properties computed without learning can serve as strong alignment targets for EEG.
The uneven gains across combinations further suggest that what a descriptor adds matters more than how many descriptors are combined.

\setcounter{figure}{1}
\begin{figure}[t]
    \centering
    \includegraphics[width=\columnwidth]{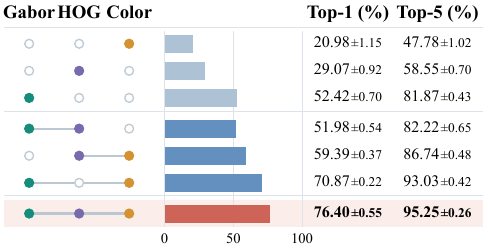}
    \caption{\textbf{Explicit visual descriptors as EEG alignment targets.} Each row is one target, and filled dots mark the descriptors it includes. Bars show 200-way within-subject top-1 accuracy.}
    \label{fig:descriptor_ablation}
\end{figure}

\subsection{Layer-wise Decodability}
\label{sec:layerwise}
\begin{figure}[t]
    \centering
    \includegraphics[width=\columnwidth]{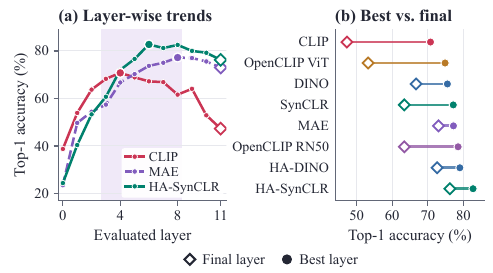}
    \caption{\textbf{Layer-wise decodability of frozen vision encoders.} (a) Top-1 accuracy of 200-way retrieval across layers for three representative encoders. The shaded band covers L03--L08, where the best layers of all eight encoders fall. (b) Best layer versus final layer for all eight encoders, where segment length shows the gap.}
    \label{fig:layerwise}
\end{figure}

Deep vision encoders build their representations progressively, from local edges and textures in early layers to increasingly semantic and task-specific features in later layers~\cite{zeiler2014visualizing,yosinski2014transferable}.
Most EEG decoding methods nonetheless align EEG with the final layer, where the machine-oriented objective acts most directly.
Is this final layer actually the best target for EEG?
To answer this, we train a separate EEG encoder for each layer of eight frozen encoders, comprising six standard pretrained models~\cite{radford2021clip,cherti2023openclip,caron2021dino,he2022mae,tian2024synclr} and two human-aligned (HA) variants of DINO and SynCLR~\cite{fu2023dreamsim,sundaram2024perceptual}.
Fig.~\ref{fig:layerwise} reveals a consistent pattern across all of them.
Accuracy peaks at an intermediate layer and then decreases overall toward the final layer, which falls 4.20 to 23.45 percentage points below the peak.
The decline is largest for CLIP, whose accuracy falls from 70.75\% at L04 to 47.30\% at its final layer, suggesting that a representation refined for a machine task is not necessarily one that EEG can decode well.
In contrast, alignment with human perception benefits EEG decoding~\cite{rajabi2025humanaligned}.
Even when all encoders are compared at their best layers, the two HA models retain a decoding advantage (Fig.~\ref{fig:layerwise}b), with HA-SynCLR at L06 reaching the highest accuracy among the intermediate layers of all eight encoders.

Taken together, the two findings point to a gap between what machine objectives optimize and what EEG can decode.
This gap suggests two ways to build better targets:
\begin{itemize}
    \item \textbf{Use what EEG decodes well.} Visual properties that EEG reliably reflects can serve as alignment targets directly, which motivates aligning EEG with fixed explicit descriptors.
    \item \textbf{Adapt the vision encoder itself.} Deep features can be adapted toward EEG instead of only being selected. Choosing the best layer of a frozen vision encoder~\cite{tang2026hcf} still searches among representations shaped by the machine objective, while the advantage of the HA models~\cite{fu2023dreamsim,sundaram2024perceptual} suggests that changing the vision encoder itself can improve decodability.
\end{itemize}
Sec.~\ref{sec:retarget} develops these two directions into the explicit visual prior branch and the EEG-anchored visual adaptation branch of ReTarget.

\section{ReTarget}
\label{sec:retarget}
\begin{figure*}[t]
    \centering
    \includegraphics[width=0.95\textwidth]{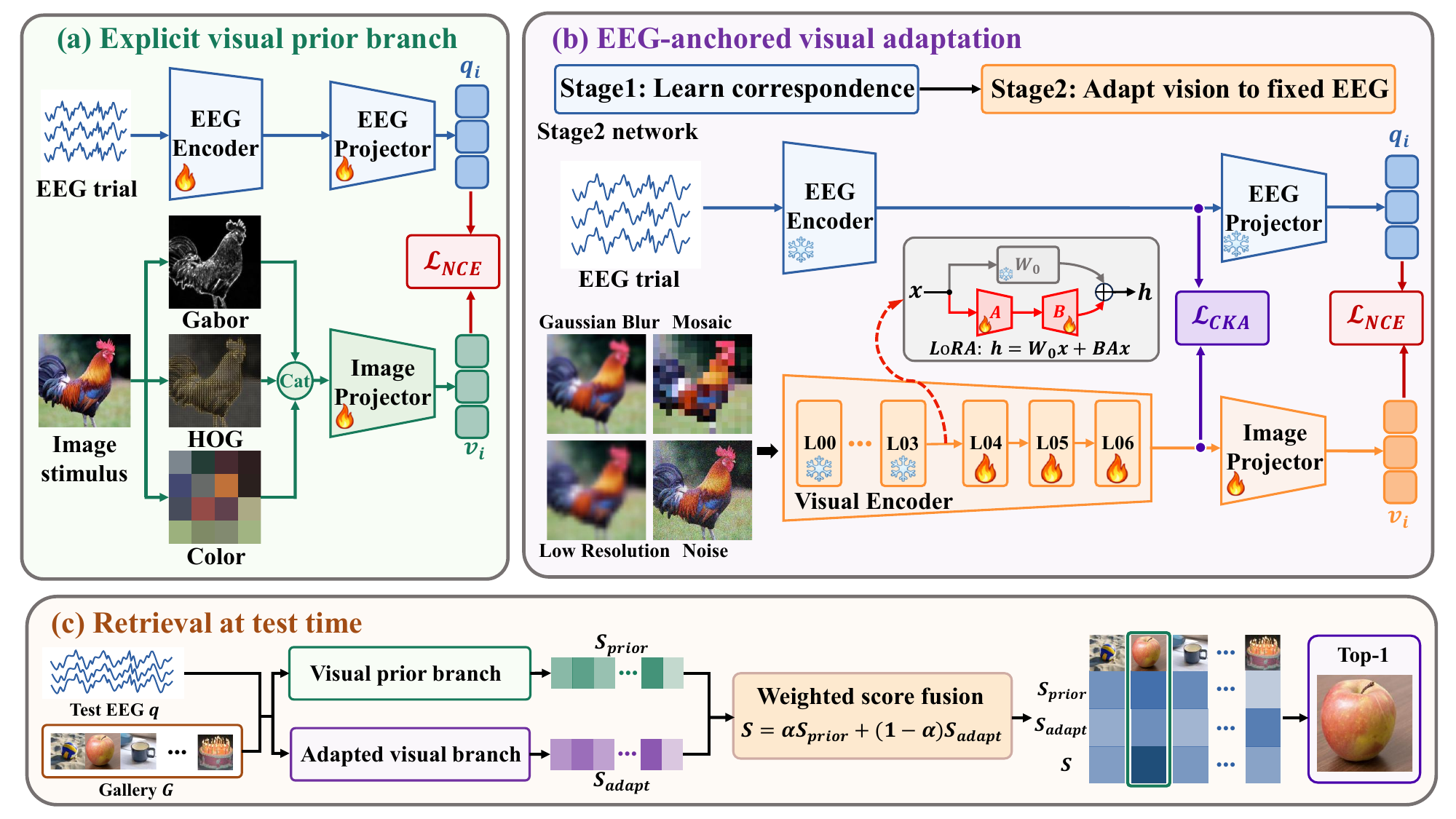}
    \caption{\textbf{Overview of ReTarget.}
    (a) The explicit visual prior branch aligns EEG with fixed, concatenated Gabor, HOG, and color descriptors.
    (b) The EEG-anchored visual adaptation branch first learns EEG--image correspondence with a frozen vision encoder (Stage~1). Stage~2 fixes the EEG encoder and its projection head and trains only LoRA in L04--L06 together with the image projection head. $\mathcal{L}_{\mathrm{CKA}}$ compares EEG and image features before projection.
    (c) At test time, a weighted sum of the two branches' scores ranks the gallery.}
    \label{fig:framework}
\end{figure*}

ReTarget consists of two retrieval branches that are trained independently (Fig.~\ref{fig:framework}).
Each branch uses the same EEG--image retrieval framework and trains its own EEG encoder and projection heads toward its own visual target $g$.
The explicit visual prior branch sets $g$ to fixed handcrafted descriptors.
The EEG-anchored visual adaptation branch sets $g$ to an intermediate layer of a deep vision encoder, which is then adapted toward a fixed EEG representation.
The branches therefore interact only at test time, where score fusion combines the cosine similarities that each branch computes over the same candidate gallery.
This design does not depend on a particular EEG encoder, since any EEG encoder trained with the same alignment objective can serve in either branch.

\subsection{Explicit Visual Prior Branch}
\label{sec:prior}
The explicit visual prior branch aligns EEG with the combination of Gabor, HOG, and color descriptors.
Gabor filters are a classical model of simple-cell receptive fields in V1~\cite{jones1987gabor} and have long been used to describe image texture~\cite{manjunath1996gabor}.
Our Gabor descriptor pools the response magnitudes of filters at multiple frequencies and orientations over a spatial grid, capturing local orientation and spatial frequency.
HOG~\cite{dalal2005hog} builds histograms of gradient orientations within local cells and thus describes the distribution of edge orientations.
The color descriptor computes per-channel color histograms within the regions of a spatial grid~\cite{swain1991color}, recording regional color statistics.
To balance the three descriptors, we reduce each to 256 dimensions with PCA fitted on training images and concatenate them into a fixed 768-d target (Sec.~1.1 of the supplementary material).

\subsection{EEG-Anchored Visual Adaptation Branch}
\label{sec:adaptation}
This branch adapts a deep vision encoder toward EEG in two stages (Fig.~\ref{fig:framework}b).
Stage~1 learns an EEG representation from a frozen visual target, and Stage~2 uses this representation as an anchor for adapting the vision encoder.

\paragraph{Stage 1: Learning the correspondence.}
Guided by Fig.~\ref{fig:layerwise}, we use HA-SynCLR ViT-B/16 as the vision encoder and take the CLS token of its L06 as the visual target.
Following NeuroBridge~\cite{zhang2026neurobridge}, 
we augment each image with Gaussian blur, Gaussian noise, low resolution, and mosaic, and average the features of the four views to obtain $u_i=g(I_i)$.
Stage~1 then follows the alignment framework of Sec.~\ref{sec:preliminaries}, training $f_E$, $P_E$, and $P_I$ with Eq.~(\ref{eq:align}) while the vision encoder stays frozen.
Stage~1 thus yields an EEG representation that already supports reasonably accurate retrieval, which makes it a reliable anchor for adaptation.

\paragraph{Stage 2: Adapting vision to fixed EEG.}
Stage~2 reverses the direction of alignment.
Instead of fitting EEG to a fixed visual target, it keeps the Stage~1 EEG encoder and projection head unchanged and moves the visual representation toward their output.
The EEG side provides two references for each sample.
The projected embedding $\bar{q}_i=P_E\big(f_E(E_i)\big)$ serves as the contrastive anchor, and the EEG feature before projection, $t_i=f_E(E_i)$, serves as a structural teacher.

We adapt the vision encoder with LoRA~\cite{hu2022lora},
which learns a low-rank update while leaving the pretrained weights intact.
For a linear projection with weight $W_0$, the adapted output is
\begin{equation}
h = W_0 x + BAx,
\label{eq:lora}
\end{equation}
where $A\in\mathbb{R}^{r\times d}$, $B\in\mathbb{R}^{d\times r}$, and $r\ll d$.
We insert LoRA into the query and value projections of the blocks leading to the readout layer (L04--L06 in Fig.~\ref{fig:framework}b) and update the image projection head $P_I$ together with it.
We denote the adapted encoder by $g_\theta$, where $\theta$ collects the LoRA parameters, and write the adapted image feature as $\tilde{u}_i=g_\theta(I_i)$.
Since $B$ is initialized to zero, $g_\theta$ coincides with the Stage~1 encoder $g$ when adaptation begins.

\paragraph{Objective.}
The two EEG references constrain the adapted features at two levels.
At the instance level, we apply Eq.~(\ref{eq:align}) with $q_i$ replaced by the anchor $\bar{q}_i$ and with $v_i=P_I(\tilde{u}_i)$, which encourages each image to retrieve its own EEG anchor.
At the relational level, samples that are similar in the EEG representation should also be similar in the visual feature space.
For a mini-batch, let $U$ and $T$ stack the adapted image features $\tilde{u}_i$ and the EEG features $t_i$, and let $U_c$ and $T_c$ denote their column-centered versions.
We measure their agreement with linear CKA~\cite{kornblith2019cka},
\begin{equation}
\mathrm{CKA}(U,T)=\frac{\|U_c^{\top}T_c\|_F^2}{\|U_c^{\top}U_c\|_F\,\|T_c^{\top}T_c\|_F}.
\label{eq:cka}
\end{equation}
Linear CKA is invariant to orthogonal transformations and isotropic scaling, so it compares the two similarity structures without requiring the features to share a coordinate system.
The Stage~2 objective is
\begin{equation}
\mathcal{L}_{\mathrm{adapt}}=\mathcal{L}_{\mathrm{NCE}}+\lambda\,\mathcal{L}_{\mathrm{CKA}},
\qquad
\mathcal{L}_{\mathrm{CKA}}=1-\mathrm{CKA}(U,T),
\label{eq:adapt}
\end{equation}
where $\lambda$ balances the two terms.
Because $\mathcal{L}_{\mathrm{CKA}}$ is computed before the projection head, it acts directly on the encoder output, whereas $\mathcal{L}_{\mathrm{NCE}}$ also updates $P_I$.

\subsection{Test-Time Score Fusion}
\label{sec:fusion}
At test time, ReTarget fuses the two branches at the score level.
For an EEG query, each branch computes cosine similarities between the query and all gallery images, and the resulting score vectors are combined as
\begin{equation}
S=\alpha\,S_{\mathrm{prior}}+(1-\alpha)\,S_{\mathrm{adapt}},
\label{eq:fusion}
\end{equation}
where $\alpha$ weights the prior branch and is selected on held-out training concepts (Sec.~\ref{sec:setup}).
This fusion adds a single hyperparameter and no trainable parameters.

\section{Experiments}

\subsection{Experimental Setup}
\label{sec:setup}
\paragraph{Dataset and protocols.}
We use THINGS-EEG~\cite{gifford2022thingseeg2}, which records EEG from ten subjects viewing natural object images. Training covers 1,654 concepts with ten images each (16,540 images), and testing uses one image from each of 200 concepts unseen during training. Following NeuroBridge~\cite{zhang2026neurobridge}, EEG responses are averaged over the repeated presentations of each image.
We also adopt its encoder configurations: the within-subject setting trains EEGProject~\cite{wu2025ubp} on 17 posterior electrodes separately for each subject, and the leave-one-subject-out (LOSO) setting trains TSConv~\cite{song2024nice} on all 63 electrodes of nine subjects and tests on the held-out one. Unless otherwise noted, inputs cover 0--500\,ms after stimulus onset at 250\,Hz. We report 200-way top-1 and top-5 accuracy averaged over subjects or LOSO folds, with $\pm$ denoting the standard deviation across five random seeds.

\paragraph{Implementation details.}
All models use AdamW~\cite{loshchilov2019adamw}, a batch size of 1,024, and 512-d projection heads. The prior branch and Stage~1 are trained for 50 epochs with a learning rate of $10^{-4}$. Stage~2 trains rank-8 LoRA for 20 epochs, with learning rates of $10^{-4}$ for LoRA and $3\times10^{-5}$ for $P_I$. To avoid checkpoint selection on test data, we report all results at the final epoch. The fusion weight $\alpha$ and the CKA weight $\lambda$ are tuned on 200 concepts held out from the training set. A grid search over $\alpha\in\{0,0.1,\dots,1\}$ on fused scores and $\lambda\in\{0,0.1,0.3,0.5\}$ on adaptation-branch scores selects $\alpha=0.4$ and $\lambda=0.3$, which we use in all experiments. Search results and further details are given in Sec.~1.2 of the supplementary material.

\subsection{Main Results}
\label{sec:main_results}
Tab.~\ref{tab:main} compares ReTarget with existing EEG-to-image retrieval methods.
ReTarget reaches 94.77\% top-1 accuracy in the within-subject setting and 30.00\% in the LOSO setting, which compares favorably with these methods in both settings.
Its individual branches show where this performance comes from.
With only handcrafted descriptors as its target, the prior branch alone already reaches 76.40\% within subjects.
The adaptation branch performs better still and exceeds HCF in both settings, indicating that adapting the vision encoder toward EEG yields a more decodable target.
Fusing the two branches adds a further 5.71 points over the adaptation branch within subjects, and the gain remains positive in the harder LOSO setting at 1.72 points.
We further examine why explicit visual priors are decodable in Sec.~\ref{sec:why_prior} and where the fusion gain comes from in Sec.~\ref{sec:complementarity}.

\newcommand{\retargetTableStdDev}[1]{{\fontsize{6.4}{8}\selectfont\textpm #1}}

\begin{table}[t]
    \retargettablefont
    \caption{\textbf{Comparison with prior methods on THINGS-EEG.}
    Top-1 and top-5 accuracy (\%) of 200-way retrieval in the within-subject and LOSO settings.
    The bottom rows report each branch of ReTarget alone and the full model with test-time score fusion.}
    \label{tab:main}
    \centering
    \fontsize{8}{8}\selectfont
    \setlength{\tabcolsep}{1pt}
    \newcommand{\mainTableMethod}[1]{{\fontsize{7.2}{8}\selectfont #1}}
    \begin{tabular*}{\columnwidth}{@{\extracolsep{\fill}}lcccc@{}}
        \toprule
        & \multicolumn{2}{c}{Within-subject} & \multicolumn{2}{c}{LOSO} \\
        \cmidrule(lr){2-3}\cmidrule(l){4-5}
        \mainTableMethod{Method} & Top-1 & Top-5 & Top-1 & Top-5 \\
        \midrule
        \mainTableMethod{NICE~\cite{song2024nice}} & 16.10 & 43.60 & 6.20 & 21.40 \\
        \mainTableMethod{ATM~\cite{li2024atm}} & 27.10 & 58.10 & 11.90 & 33.80 \\
        \mainTableMethod{Neural-MCRL~\cite{li2025neuralmcrl}} & 32.40 & 64.10 & 14.00 & 34.30 \\
        \mainTableMethod{UBP~\cite{wu2025ubp}} & 50.90 & 79.70 & 12.40 & 33.40 \\
        \mainTableMethod{NeuroBridge~\cite{zhang2026neurobridge}} & 63.20 & 89.90 & 19.00 & 45.90 \\
        \mainTableMethod{Shallow Alignment~\cite{du2026shallow}} & 82.60 & 97.70 & 21.80 & 49.40 \\
        \mainTableMethod{HCF~\cite{tang2026hcf}} & 84.60 & 98.20 & 23.40 & 54.90 \\
        \midrule
        \mainTableMethod{Prior branch} & 76.40\retargetTableStdDev{0.55} & 95.25\retargetTableStdDev{0.26} & 14.82\retargetTableStdDev{0.20} & 40.86\retargetTableStdDev{0.73} \\
        \mainTableMethod{Adaptation branch} & 89.06\retargetTableStdDev{0.20} & 98.86\retargetTableStdDev{0.14} & 28.28\retargetTableStdDev{1.05} & 59.13\retargetTableStdDev{1.92} \\
        \mainTableMethod{\textbf{ReTarget}} & \textbf{94.77}\retargetTableStdDev{0.30} & \textbf{99.73}\retargetTableStdDev{0.06} & \textbf{30.00}\retargetTableStdDev{0.97} & \textbf{60.48}\retargetTableStdDev{1.34} \\
        \bottomrule
    \end{tabular*}
\end{table}

\subsection{Why Are Explicit Visual Priors Decodable?}
\label{sec:why_prior}
\begin{figure}[t]
    \centering
    \includegraphics[width=\columnwidth]{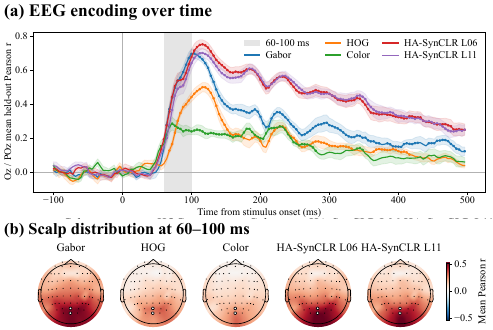}
    \caption{\textbf{Predicting EEG from visual features.} Ridge regression predicts the EEG at each electrode and time point from each visual feature, with correlations computed on held-out test images.
    (a) Mean correlation at Oz and POz over time, with shading showing $\pm1$ SEM across subjects.
    (b) Scalp maps averaged over 60--100\,ms, with white circles marking Oz and POz.}
    \label{fig:encoding}
\end{figure}

To understand why the explicit descriptors are decodable, we first use ridge regression to predict the EEG response from each visual feature.
As shown in Fig.~\ref{fig:encoding}a, Gabor features predict the EEG at Oz and POz with a correlation comparable to that of HA-SynCLR L06 within 60--100\,ms, and their group-mean curve peaks earlier, at 104\,ms against 116\,ms for both HA-SynCLR layers.
The scalp maps further show that Gabor features predict occipital electrodes more strongly than HOG and color do.
Given that early visual evoked responses involve striate and extrastriate occipital cortex~\cite{dirusso2002vep}, this pattern suggests that Gabor features capture visual attributes relevant to early cortical processing.
Across the five features, those with higher peak correlations are also more decodable as alignment targets, and HA-SynCLR L06 reaches the highest peak correlation, consistent with its high decodability in Sec.~\ref{sec:layerwise}.

\begin{figure}[t]
    \centering
    \includegraphics[width=\columnwidth]{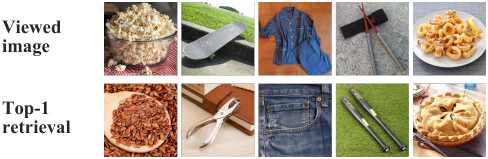}
    \caption{\textbf{Retrieval errors of the explicit visual prior branch.}
    Top: viewed images. Bottom: incorrect top-1 within-subject retrievals of the prior branch.
    In these examples, each wrongly retrieved image resembles the viewed image in a low-level visual attribute such as texture, edge orientation, or color.}
    \label{fig:error_cases}
\end{figure}

The retrieval errors of the prior branch point to the same low-level attributes.
Its wrongly retrieved images often resemble the viewed image in texture, edge orientation, or color (Fig.~\ref{fig:error_cases}), and a quantitative measure of attribute similarity further supports this observation (Sec.~3.1 of the supplementary material).

Two properties of the THINGS-EEG protocol may help explain why these low-level attributes are so decodable. First, rapid serial visual presentation (RSVP) shows each image for 100\,ms followed by a 100\,ms blank, so each new image interferes with the later processing of the previous one~\cite{gifford2022thingseeg2,mohsenzadeh2018rsvp}.
Since Gabor features predict EEG best around 100\,ms, before the next image appears at 200\,ms (Fig.~\ref{fig:encoding}a), the attributes they capture may be encoded before this interference sets in.
Second, repetition averaging more readily preserves components that are time-locked to stimulus onset and tends to attenuate temporally variable ones.
Early visual responses driven by physical image properties, such as spatial frequency and edge orientation, tend to be consistent across trials and thus gain a high signal-to-noise ratio after averaging.

\subsection{Ablation of the Adaptation Branch}
\label{sec:adapt_ablation}
To examine how each part of the adaptation branch contributes to retrieval, we ablate the trainable components of Stage~2 (Tab.~\ref{tab:adapt_ablation}).
Starting from the frozen encoder, updating only the image projection head raises top-1 accuracy from 85.70\% to 86.15\%, whereas updating only LoRA raises it to 87.40\%, suggesting that adapting the visual encoding itself contributes more than refitting the projection head.
Updating both yields a further gain.
The contrastive loss, however, only aligns each image with its own EEG anchor and does not constrain how samples relate to one another.
We therefore add the CKA loss, which matches the geometry of the image features to that of the fixed EEG features, so that images evoking similar EEG responses also lie close in the visual feature space.
With this relational constraint, the adaptation branch reaches 89.06\% and gives the highest top-1 accuracy both alone and after fusion.
The same adaptation procedure also improves retrieval on backbones with other pretraining objectives, such as DINO and MAE, though HA-SynCLR remains the strongest target (Sec.~2.1 of the supplementary material).

\begin{table}[t]
    \retargettablefont
    \caption{\textbf{Ablation of trainable components in Stage~2 of the adaptation branch.}
    Within-subject top-1 and top-5 accuracy (\%) of the adaptation branch alone and after score fusion.
    All variants start from the same Stage~1 model, and \textit{Frozen encoder} skips Stage~2.}
    \label{tab:adapt_ablation}
    \centering
    \footnotesize
    \setlength{\tabcolsep}{2pt}
    \begin{tabular*}{\columnwidth}{@{\extracolsep{\fill}}lcccc@{}}
        \toprule
        & \multicolumn{2}{c}{Adaptation branch} & \multicolumn{2}{c}{After fusion} \\
        \cmidrule(lr){2-3}\cmidrule(l){4-5}
        Stage~2 variant & Top-1 & Top-5 & Top-1 & Top-5 \\
        \midrule
        Frozen encoder & 85.70 & 98.05 & 93.50 & 99.60 \\
        Projection head only & 86.15 & 98.20 & 93.90 & 99.60 \\
        LoRA only & 87.40 & 98.35 & 94.40 & 99.70 \\
        LoRA + projection head & 88.15 & 98.65 & 94.40 & \textbf{99.75} \\
        LoRA + projection head + CKA & \textbf{89.06} & \textbf{98.75} & \textbf{94.77} & 99.70 \\
        \bottomrule
    \end{tabular*}
\end{table}

\begin{figure}[t]
    \centering
    \includegraphics[width=0.76\columnwidth]{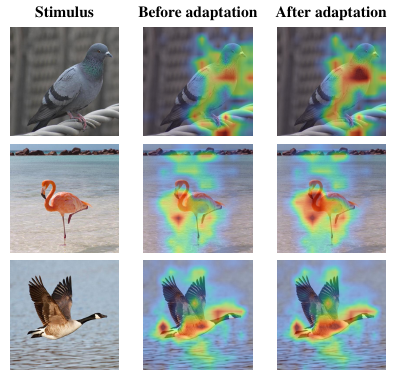}
    \caption{\textbf{Grad-CAM~\cite{selvaraju2017gradcam} maps before and after visual adaptation.}
    Maps are computed at L06 of the frozen and adapted HA-SynCLR for the similarity between each test image and its fixed Stage~1 EEG representation, and the two maps in each row share one color scale.}
    \label{fig:gradcam}
\end{figure}

Fig.~\ref{fig:gradcam} visualizes which image regions contribute to the EEG--image similarity before and after adaptation.
In the examples shown, adaptation strengthens the contribution of object parts, such as the chest of the pigeon and the lower neck of the flamingo.

\subsection{Generalization across EEG Encoders}
\label{sec:encoders}
Because the adaptation branch uses the fixed EEG representation as its anchor, its benefit may depend on the EEG encoder that produces this anchor.
To test whether this benefit generalizes beyond EEGProject, we train both branches with four other EEG encoders: TSConv~\cite{song2024nice}, ATM-S~\cite{li2024atm}, EEGNet~\cite{lawhern2018eegnet}, and EEGTransformer (Tab.~\ref{tab:encoders}).
Adaptation improves top-1 accuracy over the frozen visual target for all five encoders, by 3.36 to 18.25 points, and fusion with the prior branch improves accuracy further in every case.
The benefit of ReTarget is therefore not tied to a particular EEG architecture.

\begin{table}[t]
    \retargettablefont
    \caption{\textbf{Retrieval with different EEG encoders.}
    Within-subject top-1 accuracy (\%) with both branches trained for each EEG encoder.
    \textit{Frozen} and \textit{Adapted} denote the visual target before and after adaptation, and \textit{Fusion} combines the adapted branch with the prior branch.}
    \label{tab:encoders}
    \centering
    \small
    \setlength{\tabcolsep}{2pt}
    \begin{tabular*}{\columnwidth}{@{\extracolsep{\fill}}lccc@{}}
        \toprule
        EEG encoder & Frozen & Adapted & Fusion \\
        \midrule
        EEGProject     & 85.70 & 89.06\,{\scriptsize (+3.36)} & \textbf{94.77}\,{\scriptsize (+5.71)} \\
        TSConv         & 72.45 & 82.20\,{\scriptsize (+9.75)} & \textbf{88.25}\,{\scriptsize (+6.05)} \\
        ATM-S          & 80.15 & 87.70\,{\scriptsize (+7.55)} & \textbf{91.60}\,{\scriptsize (+3.90)} \\
        EEGNet         & 47.00 & 65.25\,{\scriptsize (+18.25)} & \textbf{72.10}\,{\scriptsize (+6.85)} \\
        EEGTransformer & 66.00 & 77.00\,{\scriptsize (+11.00)} & \textbf{87.10}\,{\scriptsize (+10.10)} \\
        \bottomrule
    \end{tabular*}
\end{table}

\begin{figure*}[t]
    \centering
    \includegraphics[width=0.92\textwidth]{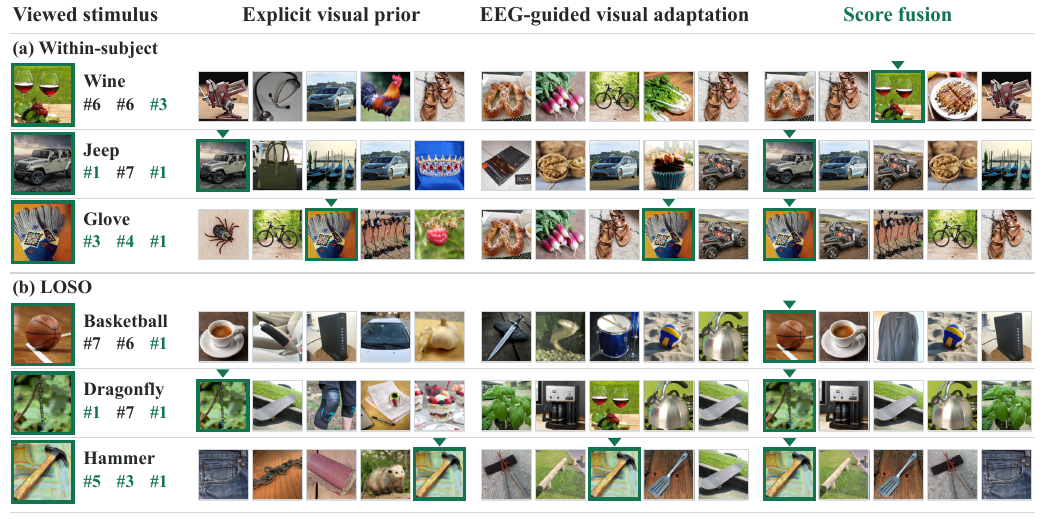}
    \caption{\textbf{Retrieval cases of the two branches and their fusion.}
    (a) Within-subject and (b) LOSO retrieval.
    Each group shows the top-5 retrievals, and green boxes mark the correct image.
    For each stimulus, the three ranks give the position of the correct image among all 200 candidates for the prior branch, the adaptation branch, and fusion.}
    \label{fig:fusion_cases}
\end{figure*}

\subsection{Complementarity of Visual Targets}
\label{sec:complementarity}
To analyze where the fusion gain comes from, we align separate EEG models to different visual targets and fuse them in pairs with the same test-time score fusion (Tab.~\ref{tab:complementarity}).
Two models aligned to the same HA-SynCLR L06 target gain only 1.33 points, which provides a reference for the benefit of model count alone.
Replacing one of them with another layer of HA-SynCLR or with OpenCLIP-RN50 moderately increases the gain, suggesting that different deep representations carry partly different information.
Interestingly, pairing the explicit visual prior with either deep target yields far larger gains of 8.13 and 11.05 points, even though the prior alone is weaker than both deep targets.
This suggests that the explicit visual prior is strongly complementary to deep pretrained representations.
The fusion gain of ReTarget thus arises primarily from the complementarity of its visual targets rather than from the number of models.

\begin{table}[t]
    \retargettablefont
    \caption{\textbf{Fusion gains of different visual-target pairs.}
    Within-subject top-1 accuracy (\%) for two EEG models aligned to frozen targets A and B and their fusion. Gain is over the stronger model. The first row uses the same target with different seeds. L06/L08 and RN50 L03 denote HA-SynCLR and OpenCLIP-RN50 layers, respectively. Prior denotes the explicit visual prior.}
    \label{tab:complementarity}
    \centering
    \fontsize{8}{8}\selectfont
    \setlength{\tabcolsep}{1.5pt}
    \begin{tabular*}{\columnwidth}{@{\extracolsep{\fill}}lcccc@{}}
        \toprule
        Target pair & A & B & Fusion & Gain (pp) \\
        \midrule
        L06 + L06 & 85.35\retargetTableStdDev{0.49} & 85.35\retargetTableStdDev{0.49} & 86.68\retargetTableStdDev{0.14} & +1.33 \\
        L06 + L08 & 85.35\retargetTableStdDev{0.49} & 84.76\retargetTableStdDev{0.21} & 88.68\retargetTableStdDev{0.30} & +3.33 \\
        L06 + RN50 L03 & 85.35\retargetTableStdDev{0.49} & 81.66\retargetTableStdDev{0.85} & 88.95\retargetTableStdDev{0.55} & +3.60 \\
        \midrule
        Prior + L06 & 76.40\retargetTableStdDev{0.55} & 85.35\retargetTableStdDev{0.49} & 93.48\retargetTableStdDev{0.44} & +8.13 \\
        Prior + RN50 L03 & 76.40\retargetTableStdDev{0.55} & 81.66\retargetTableStdDev{0.85} & 92.71\retargetTableStdDev{0.47} & +11.05 \\
        \bottomrule
    \end{tabular*}
\end{table}

\subsection{Error Correction by Fusion}
\label{sec:correction}
To illustrate how the complementarity of the two targets corrects individual retrieval errors, we visualize retrieval cases of each branch and of their fusion in Fig.~\ref{fig:fusion_cases}.
When one branch retrieves the correct image, fusion tends to preserve this answer, as for the jeep that the prior branch ranks first but the adaptation branch misses.
Fusion can also correct errors made by both branches.
For example, neither branch places the wine within subjects or the basketball in the LOSO setting in its top five, yet fusion moves them into the top five and to the first place, respectively.
These corrections are not limited to the examples shown.
Across all within-subject test queries, fusion corrects 697 top-1 errors of the adaptation branch while introducing only 126 new ones, and it recovers 143 of the 428 queries that both branches miss (Sec.~3.2 of the supplementary material).
Because fusion ranks candidates by the combined scores of both branches, a correct image scored highly by both can surpass candidates favored by only one, allowing fusion to recover it from two incomplete rankings.

\section{Conclusion}

This paper revisits the visual target that EEG is aligned to in EEG-to-image retrieval.
We show that representations optimized for machine tasks are not necessarily the most decodable for EEG, as handcrafted low-level descriptors alone already support strong retrieval.
Building on this observation, ReTarget combines an explicit visual prior with EEG-anchored visual adaptation, which adapts the vision encoder toward a fixed EEG representation instead of keeping it frozen.
ReTarget reaches 94.77\% within-subject top-1 accuracy on THINGS-EEG, with gains driven by complementary targets rather than more models.
These results suggest that the visual target should be designed for what EEG can reliably reflect and can be actively adapted toward EEG rather than fixed in advance.
However, the explicit visual prior is less effective across subjects.
Improving cross-subject robustness and reducing the reliance on repetition averaging are promising directions for real-world use.

\section*{LLM Usage Statement}

We used OpenAI ChatGPT and Anthropic Claude to assist with manuscript drafting, language refinement, and literature organization. OpenAI Codex was used to assist with LaTeX editing, code development, and figure and table preparation.  All AI-assisted content was reviewed and revised by the authors, who take full responsibility for the research design, experimental results, references, and scientific claims presented in this paper.

{\small
\bibliographystyle{ieeenat_fullname}
\bibliography{references}
}

\onecolumn
\twocolumn[{\centering\Large\bfseries Supplementary Material\par\vspace{1em}}]
\setcounter{section}{0}
\renewcommand{\thesection}{\arabic{section}}
\renewcommand{\thesubsection}{\thesection.\arabic{subsection}}

\setcounter{table}{0}
\setcounter{figure}{0}
\setcounter{equation}{0}
\renewcommand{\thetable}{S\arabic{table}}
\renewcommand{\thefigure}{S\arabic{figure}}
\renewcommand{\theequation}{S\arabic{equation}}

\renewcommand{\theHsection}{supp.\arabic{section}}
\renewcommand{\theHtable}{supp.\arabic{table}}
\renewcommand{\theHfigure}{supp.\arabic{figure}}
\renewcommand{\theHequation}{supp.\arabic{equation}}
\captionsetup[table]{font={small,retargettimes}}

\section{Implementation Details}
\label{sec:supp-implementation}

\subsection{Explicit Visual Descriptors}
\label{sec:supp-descriptors}
\raggedbottom
The explicit visual prior branch uses three handcrafted descriptors, whose parameters are listed in Tab.~\ref{tab:supp-descriptor-parameters}.
All descriptors are computed on the original image resized to $224\times224$, without the four augmented views used by the adaptation branch.
The Gabor descriptor filters the grayscale image with a bank of Gabor filters at several frequencies and orientations.
It then summarizes the response magnitude of each filter by its mean and standard deviation within each region of a spatial grid.
HOG builds histograms of gradient orientations over local cells of the grayscale image and normalizes them within overlapping blocks of cells.
The color descriptor computes a histogram for each RGB channel within each region of a spatial grid and normalizes each histogram to sum to one.

The three descriptors differ in raw dimensionality and numerical range.
To keep any one of them from dominating the target, we standardize each descriptor per dimension and reduce it to 256 dimensions with PCA.
Both steps are fitted on the training images and applied unchanged to the test images.
Concatenating the three 256-d vectors gives the 768-d target of the prior branch.

For a controlled comparison, all seven targets in Fig.~2 of the main paper have the same dimensionality of 768.
In each pair, every descriptor is standardized in the same way and reduced to 384 dimensions with PCA.
The three-descriptor combination is the target described above.
Among the single descriptors, Gabor and color keep their original 768-d representations, and PCA reduces HOG from 1,296 to 768 dimensions.

\begin{table}[!ht]
    \retargettablefont
    \caption{\textbf{Parameters of the explicit visual descriptors.}
    Dim.\ gives the dimensionality of each descriptor before and after PCA in the 768-d target of the prior branch.}
    \label{tab:supp-descriptor-parameters}
    \centering
    \normalsize
    \setlength{\tabcolsep}{4pt}
    \renewcommand{\arraystretch}{1.15}
    \begin{tabular*}{\columnwidth}{@{\extracolsep{\fill}}lp{0.48\columnwidth}c@{}}
        \toprule
        Descriptor & Parameters & Dim. \\
        \midrule
        Gabor & \raggedright Grayscale; frequencies 0.1, 0.2, 0.3, 0.4 cycles/pixel; 6 orientations evenly spaced over 180°; mean and std of magnitude in 4×4 regions & 768 → 256 \\
        \addlinespace
        HOG & \raggedright Grayscale; 9 orientation bins; 32×32-pixel cells; 2×2-cell blocks & 1,296 → 256 \\
        \addlinespace
        Color & \raggedright RGB; 16 bins per channel; 4×4 regions; each histogram normalized to sum to one & 768 → 256 \\
        \bottomrule
    \end{tabular*}
\end{table}

\subsection{Hyperparameter Selection on Held-out Training Concepts}
\label{sec:supp-hyperparameters}
We select the fusion weight $\alpha$ and the CKA weight $\lambda$ on a validation set built from the training data.
We hold out 200 of the 1,654 training concepts and train all models on the remaining 1,454 concepts.
The 2,000 held-out images are evaluated with the same 200-way retrieval protocol as the test set.
EEG whitening, descriptor standardization, and PCA are estimated on the training concepts only.
All models are trained in the within-subject setting with the main training schedule and evaluated at the final epoch.
Each hyperparameter is chosen by its validation top-1 accuracy.
Because validation EEG is averaged over 4 repetitions rather than 80, validation accuracies are lower than test accuracies.

Because $\lambda$ affects only the adaptation branch, we compare $\lambda\in\{0,0.1,0.3,0.5\}$ on the scores of this branch alone.
Removing the CKA loss ($\lambda=0$) gives the lowest accuracy, and $\lambda=0.3$ gives the highest in both top-1 and top-5 (Tab.~\ref{tab:supp-lambda-selection}).
We therefore fix $\lambda=0.3$ in all subsequent experiments.

\begin{figure}[t]
    \centering
    \includegraphics[width=\columnwidth]{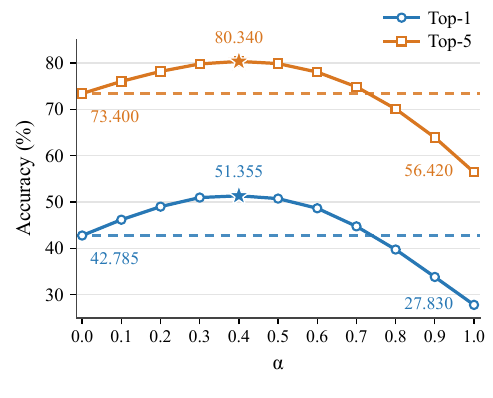}
    \caption{\textbf{Selection of the fusion weight $\alpha$ on held-out training concepts.}
    Fused retrieval accuracy with $\lambda=0.3$ on the same validation queries as Tab.~\ref{tab:supp-lambda-selection}.
    $\alpha=0$ and $\alpha=1$ correspond to the adaptation branch and the prior branch alone.
    Dashed lines mark the accuracy of the adaptation branch, and the star marks the selected value.}
    \label{fig:supp-alpha-selection}
\end{figure}

We sweep $\alpha$ from 0 to 1 in steps of 0.1 on the fused scores.
Ties would be broken in favor of the value closest to 0.5 and then the smaller value.
Fused accuracy rises from both single-branch endpoints and peaks at $\alpha=0.4$ in both top-1 and top-5 (Fig.~\ref{fig:supp-alpha-selection}).
Every weight from 0.1 to 0.7 outperforms the stronger single branch, so the fusion gain does not hinge on a precisely tuned $\alpha$.
Based on these results, we use $\alpha=0.4$ in all experiments.
Both $\alpha$ and $\lambda$ are shared by the within-subject and LOSO settings without further tuning for individual subjects or seeds.

\begin{table}[t]
    \retargettablefont
    \caption{\textbf{Selection of the CKA weight λ on held-out training concepts.}
    Retrieval accuracy of the adaptation branch alone over 20,000 validation queries from ten subjects.
    \# is the number of correct queries.
    λ = 0 removes the CKA loss.
    Bold marks the selected value.}
    \label{tab:supp-lambda-selection}
    \centering
    \normalsize
    \setlength{\tabcolsep}{3pt}
    \renewcommand{\arraystretch}{1.15}
    \begin{tabular*}{\columnwidth}{@{\extracolsep{\fill}}ccccc@{}}
        \toprule
        λ & Top-1 \# & Top-1 (\%) & Top-5 \# & Top-5 (\%) \\
        \midrule
        0.0 & 8,454 & 42.270 & 14,605 & 73.025 \\
        0.1 & 8,498 & 42.490 & 14,642 & 73.210 \\
        \textbf{0.3} & \textbf{8,557} & \textbf{42.785} & \textbf{14,680} & \textbf{73.400} \\
        0.5 & 8,542 & 42.710 & 14,662 & 73.310 \\
        \bottomrule
    \end{tabular*}
\end{table}

\subsection{Additional Implementation Details}
\label{sec:supp-additional-implementation}
The within-subject setting uses 17 posterior electrodes (P7, P5, P3, P1, Pz, P2, P4, P6, P8, PO7, PO3, POz, PO4, PO8, O1, Oz, and O2), while the LOSO setting uses all 63 electrodes.
For the adaptation branch, the four augmented views of each $224\times224$ image are generated by Gaussian blur ($15\times15$ kernel), Gaussian noise (standard deviation 25 on the 0--255 pixel scale), low resolution (downsampling to 10\% of the image size and upsampling back), and mosaic (blocks of about $5\times5$ pixels).
Following NeuroBridge~\cite{zhang2026neurobridge}, we fix the temperature $\tau$ in Eq.~(1) at 0.07.
During adaptation in the LOSO setting, each batch contains EEG from a single source subject, which prevents anchors of the same image from other subjects from being treated as negatives.

\section{Deep Visual Targets and EEG-Anchored Adaptation}
\label{sec:supp-visual-targets}

\begin{figure*}[!tp]
    \begin{minipage}[t][\textheight][t]{\textwidth}
    \centering
    \begin{subfigure}{\textwidth}
        \centering
        \includegraphics[width=\linewidth]{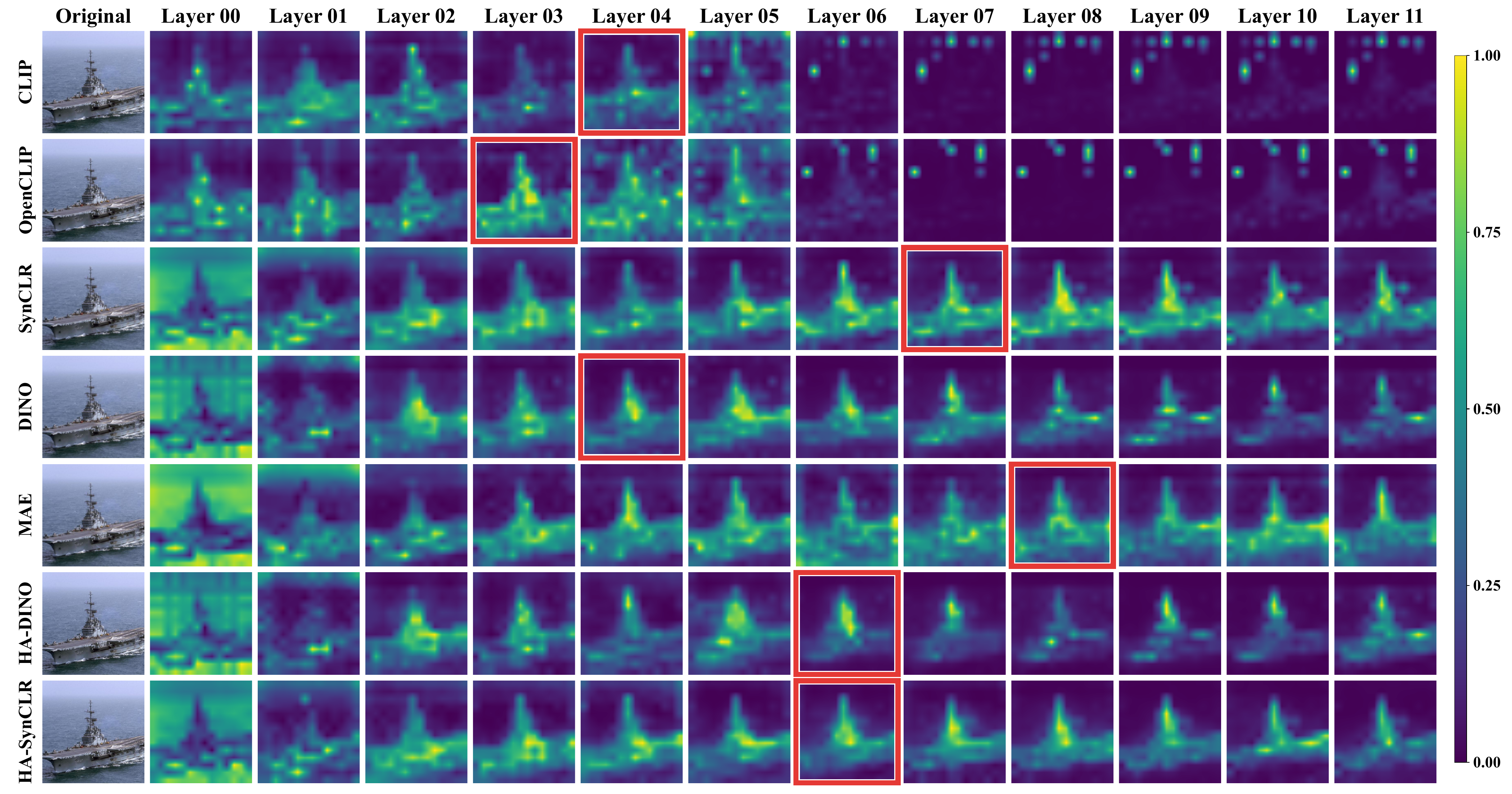}
        \caption{Aircraft carrier}
        \label{fig:supp-layerwise-attention-ship}
    \end{subfigure}

    \vspace{0.5em}

    \begin{subfigure}{\textwidth}
        \centering
        \includegraphics[width=\linewidth]{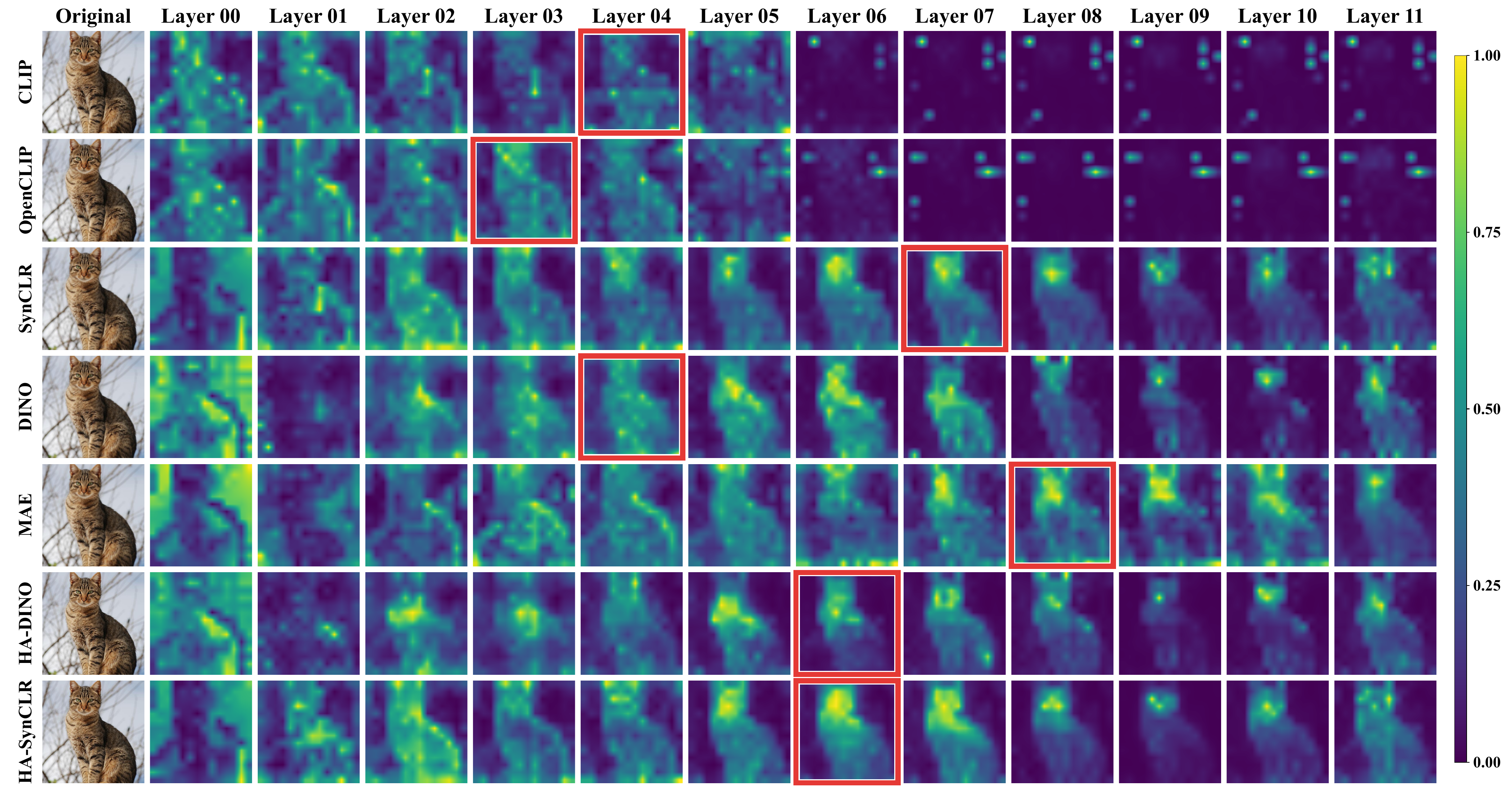}
        \caption{Cat}
        \label{fig:supp-layerwise-attention-cat}
    \end{subfigure}
    \caption{\textbf{Layer-wise CLS-to-patch attention of seven frozen vision encoders for two THINGS-EEG test images:} (a) aircraft carrier and (b) cat.
    Red boxes mark the best decoding layer of each encoder in Fig.~3 of the main paper.
    Each map is min--max normalized within its panel and bilinearly interpolated, and the color bar shows this relative attention.}
    \label{fig:supp-layerwise-attention}
    \end{minipage}
\end{figure*}

\subsection{Adaptation on Other ViT Backbones}
\label{sec:supp-other-backbones}
To evaluate how well EEG-anchored visual adaptation generalizes across vision encoders, we apply it to five other ViT backbones with different pretraining objectives.
We keep the procedure of the main paper and change only the vision encoder.
For each backbone, Stage~1 trains the EEG encoder and projection heads on the frozen features of its best layer in Fig.~3 of the main paper.
Stage~2 then fixes the EEG side and trains LoRA together with the image projection head.
LoRA is inserted into the query and value projections of the readout block and the two blocks before it.
All other settings, including the CKA loss and the fusion weight, follow the main configuration.

As shown in Tab.~\ref{tab:supp-other-backbones}, EEG-anchored visual adaptation improves top-1 accuracy on all six backbones, both alone and after fusion with the prior branch.
Among these backbones, HA-SynCLR remains the best vision encoder.
It has the highest frozen accuracy and keeps the lead after adaptation and fusion, which supports its choice as the visual target of ReTarget.

\begin{table*}[t]
    \retargettablefont
    \caption{\textbf{Adaptation on other ViT backbones.}
    Within-subject top-1 accuracy (\%).
    Readout is the layer used as the visual target for each backbone.
    Δ is the gain of Adapted over Frozen, and + Prior denotes fusion with the prior branch.}
    \label{tab:supp-other-backbones}
    \centering
    \normalsize
    \setlength{\tabcolsep}{2pt}
    \renewcommand{\arraystretch}{1.15}
    \begin{tabular*}{\textwidth}{@{\extracolsep{\fill}}llcccccc@{}}
        \toprule
        Pretraining & Backbone & Readout & Frozen & Adapted & Δ & Frozen + Prior & Adapted + Prior \\
        \midrule
        Image–text & OpenCLIP ViT-B/16 & L03 & 76.35 & 79.35 & +3.00 & 90.65 & 91.55 \\
        \midrule
        Self-supervised & DINO & L04 & 79.05 & 83.40 & +4.35 & 91.25 & 92.10 \\
        Self-supervised & MAE & L08 & 81.70 & 84.60 & +2.90 & 91.95 & 92.50 \\
        Self-supervised & SynCLR & L07 & 80.85 & 84.60 & +3.75 & 91.95 & 92.60 \\
        \midrule
        Human-aligned & HA-DINO & L06 & 82.90 & 84.95 & +2.05 & 92.10 & 92.50 \\
        \textbf{Human-aligned} & \textbf{HA-SynCLR} & \textbf{L06} & \textbf{85.70} & \textbf{89.06} & \textbf{+3.36} & \textbf{93.50} & \textbf{94.77} \\
        \bottomrule
    \end{tabular*}
\end{table*}

\subsection{Layer-wise Attention of Vision Encoders}
\label{sec:supp-layerwise}
Representations in deep vision encoders change with depth, from local edges and textures in early layers to more abstract and task-specific features in later ones~\cite{zeiler2014visualizing,yosinski2014transferable}.
Fig.~3 of the main paper shows that decodability also varies with depth, peaking at an intermediate layer and declining in deeper layers.
To see how the image content attended by each layer changes, Fig.~\ref{fig:supp-layerwise-attention} visualizes the attention of the CLS token, which ReTarget uses as its visual target, in each layer of seven frozen ViT-B/16 encoders.
In each layer, we take the attention weights from the CLS token to the 196 image patches, average them over the 12 heads, and arrange them on the $14\times14$ patch grid.

Taking HA-SynCLR as an example, the attention changes with depth in three stages.
In shallow layers (L00--L02), it is dispersed and responds to local edges and textures.
In middle layers (L04--L07), it converges on the object, and the overall shape of the ship and the posture of the cat become clearly visible.
In deeper layers (L09--L11), it concentrates on a few local regions, such as the superstructure of the ship and the head of the cat, and the outline of the object becomes less continuous.
SynCLR, DINO, MAE, and HA-DINO follow a broadly similar progression.

CLIP and OpenCLIP show a different pattern in deeper layers.
From L06 on, their strongest attention moves from the object to a few isolated patches, mostly in the background.
This concentration is present in the raw attention weights rather than produced by the display normalization.
At L00, the five most attended patches receive only 4--5\% of the patch attention, close to the 2.6\% expected under uniform attention.
At L11, they receive 35--52\%.
In other words, a handful of the 196 patches take up more than a third of the CLS attention in the final layer.
This pattern resembles the artifacts reported by Darcet et al.~\cite{darcet2024registers}, who observed that attention in several ViTs, including OpenCLIP, peaks on a few tokens in low-information background regions and suggested that these tokens are repurposed for internal computation.
If the same holds here, the CLS token in these layers may draw less on the local structure of the object.
This is consistent with the steeper decline in decodability of CLIP and OpenCLIP in deeper layers (Fig.~3).

\subsection{Position of the CKA Loss}
\label{sec:supp-cka-position}
The CKA loss can be computed on the features before or after the projection head on each side, which gives four possible combinations.
Features before projection are 768-d, and those after projection are 512-d.
We compare the four combinations in Tab.~\ref{tab:supp-cka-position} with all other settings fixed.
Both the adaptation branch and the fused model change little across the four positions, so the position of the CKA loss is not critical to adaptation.
We therefore keep CKA before projection on both sides, where the loss acts directly on the output of the vision encoder (Sec.~4.2 of the main paper).

\begin{table}[t]
    \retargettablefont
    \caption{\textbf{Position of the CKA loss.}
    Within-subject top-1 and top-5 accuracy (\%) of the adaptation branch alone and after score fusion.
    Before and after refer to the features before and after the projection head on each side.
    The first row is the default setting.}
    \label{tab:supp-cka-position}
    \centering
    \normalsize
    \setlength{\tabcolsep}{3pt}
    \renewcommand{\arraystretch}{1.15}
    \begin{tabular*}{\columnwidth}{@{\extracolsep{\fill}}cccccc@{}}
        \toprule
        \multicolumn{2}{@{}c}{CKA Position} & \multicolumn{2}{c}{Adaptation} & \multicolumn{2}{c@{}}{Fusion} \\
        \cmidrule(lr){1-2}\cmidrule(lr){3-4}\cmidrule(l){5-6}
        Image & EEG & Top-1 & Top-5 & Top-1 & Top-5 \\
        \midrule
        Before & Before & 89.06 & 98.86 & 94.77 & 99.73 \\
        Before & After & 89.20 & 98.75 & 94.65 & 99.70 \\
        After & Before & 89.05 & 98.75 & 94.80 & 99.70 \\
        After & After & 88.75 & 98.75 & 94.70 & 99.75 \\
        \bottomrule
    \end{tabular*}
\end{table}

\subsection{Full Results with Different EEG Encoders}
\label{sec:supp-eeg-encoders}
Tab.~3 of the main paper shows that adaptation improves retrieval for five EEG encoders, and Tab.~\ref{tab:supp-eeg-encoders} gives the full results.
For every encoder, fusing the prior branch with the adapted target outperforms fusing it with the frozen target.
This indicates that adaptation remains effective even after fusion.
The improvement tends to be larger for encoders with lower frozen accuracy, ranging from 1.27 points in top-1 accuracy for EEGProject to 11.80 points for EEGNet.

The five EEG encoders span MLP, convolutional, and Transformer architectures.
EEGProject, TSConv, ATM-S, and EEGNet are adapted from prior work~\cite{wu2025ubp,song2024nice,li2024atm,lawhern2018eegnet}, and EEGTransformer uses the implementation in the NeuroBridge codebase~\cite{zhang2026neurobridge}.
EEGTransformer treats the EEG at each time point as a token, projects its 17 channels to 128 dimensions, and adds sinusoidal positional encodings.
Four Transformer encoder layers with eight attention heads then process the token sequence.
The output tokens are averaged over time and linearly projected to the EEG feature.

\begin{table*}[t]
    \retargettablefont
    \caption{\textbf{Full within-subject results with different EEG encoders.}
    Each cell gives top-1 / top-5 accuracy (\%).
    Frozen and Adapted denote the adaptation branch before and after Stage~2, and Prior denotes the prior branch alone.
    Frozen + Prior and ReTarget fuse the prior branch with the frozen and adapted branches.}
    \label{tab:supp-eeg-encoders}
    \centering
    \normalsize
    \setlength{\tabcolsep}{4pt}
    \renewcommand{\arraystretch}{1.15}
    \begin{tabular*}{\textwidth}{@{\extracolsep{\fill}}lccccc@{}}
        \toprule
        EEG encoder & Frozen & Adapted & Prior & Frozen + Prior & ReTarget \\
        \midrule
        EEGProject & 85.70 / 98.05 & 89.06 / 98.86 & 76.40 / 95.25 & 93.50 / 99.60 & 94.77 / 99.73 \\
        TSConv & 72.45 / 94.10 & 82.20 / 97.20 & 59.60 / 88.05 & 82.70 / 97.90 & 88.25 / 98.60 \\
        ATM-S & 80.15 / 96.85 & 87.70 / 98.35 & 66.90 / 91.60 & 86.75 / 98.80 & 91.60 / 99.50 \\
        EEGNet & 47.00 / 79.80 & 65.25 / 93.20 & 40.45 / 76.20 & 60.30 / 89.10 & 72.10 / 95.35 \\
        EEGTransformer & 66.00 / 92.90 & 77.00 / 96.80 & 66.45 / 90.75 & 82.50 / 97.75 & 87.10 / 98.75 \\
        \bottomrule
    \end{tabular*}
\end{table*}

\section{Analyses of the Explicit Prior and Fusion}
\label{sec:supp-prior-fusion}

\subsection{Attribute Similarity of Retrieval Errors}
\label{sec:supp-error-attributes}
Fig.~6 of the main paper shows that the wrong retrievals of the prior branch often resemble the viewed image in a low-level attribute.
To quantify this observation, we measure how similar each wrongly retrieved image is to the correct image in terms of Gabor, HOG, and color descriptors.
For each top-1 error, we rank the other 199 candidates by their similarity to the correct image under one attribute and take the percentile of the retrieved image.
A high percentile means that the error lies among the candidates most similar to the correct image.
However, a high percentile could also arise if the wrongly retrieved images are similar to many images in general.
We therefore build a baseline that randomly re-pairs the wrong images with queries while keeping how often each image is retrieved.
Tab.~\ref{tab:supp-error-attributes} reports how far the actual percentile exceeds this baseline, which reflects similarity specific to the correct image.

All entries are positive, so the errors of every target resemble the correct image more than the random baseline does.
Each single-descriptor model shows its largest increase on its own attribute.
The Gabor and HOG models also show large increases on each other's attribute, since both descriptors encode local orientation.
In contrast, the increases between color and either Gabor or HOG remain small.
This separation is consistent with Fig.~2 of the main paper, where color complements Gabor far more than HOG does.
Models trained with combined targets make errors that resemble the correct image in several attributes at once, and the full prior branch follows the same pattern.

\begin{table}[t]
    \retargettablefont
    \caption{\textbf{Attribute similarity of top-1 retrieval errors.}
    Rows are the training targets, and columns are the attributes used to measure similarity.
    Each entry is the increase in the similarity percentile of the wrongly retrieved image over random re-pairing, in percentile points.
    Bold marks the largest entry of each single-descriptor model.}
    \label{tab:supp-error-attributes}
    \centering
    \normalsize
    \setlength{\tabcolsep}{3pt}
    \renewcommand{\arraystretch}{1.15}
    \begin{tabular*}{\columnwidth}{@{\extracolsep{\fill}}lccc@{}}
        \toprule
        Training target & Gabor & HOG & Color \\
        \midrule
        Gabor & \textbf{+29.66} & +17.61 & +4.32 \\
        HOG & +19.95 & \textbf{+21.05} & +2.64 \\
        Color & +4.32 & +4.96 & \textbf{+27.57} \\
        \midrule
        Gabor + HOG & +27.37 & +19.33 & +5.39 \\
        Gabor + Color & +25.57 & +14.45 & +20.61 \\
        HOG + Color & +17.18 & +17.43 & +20.37 \\
        Gabor + HOG + Color & +24.84 & +16.50 & +17.43 \\
        \bottomrule
    \end{tabular*}
\end{table}

\subsection{Error Transitions under Score Fusion}
\label{sec:supp-error-transitions}
Sec.~5.7 of the main paper reports how fusion changes the top-1 errors of the adaptation branch.
Tab.~\ref{tab:supp-error-transitions} gives the complete within-subject statistics, pooled over the 2,000 test queries and five random seeds (10,000 evaluations).
It groups the evaluations by whether each branch alone retrieves the correct image at top-1 and counts how many of them the fused model answers correctly.

Fusion fixes far more errors than it introduces.
Of the 1,094 errors of the adaptation branch, fusion corrects 697 (63.71\%).
These fixes come from two cases.
In 554 of them, only the prior branch is correct, and fusion adopts its answer.
In the other 143, both branches are wrong, yet the combined scores still place the correct image first.
Fusion breaks only 126 correct answers of the adaptation branch.
These all occur when the prior branch is wrong, as fusion keeps every answer that both branches get right.
The net gain is therefore 5.71 points.

\begin{figure*}[!tp]
    \begin{minipage}[t][\textheight][t]{\textwidth}
    \centering
    \includegraphics[width=\linewidth]{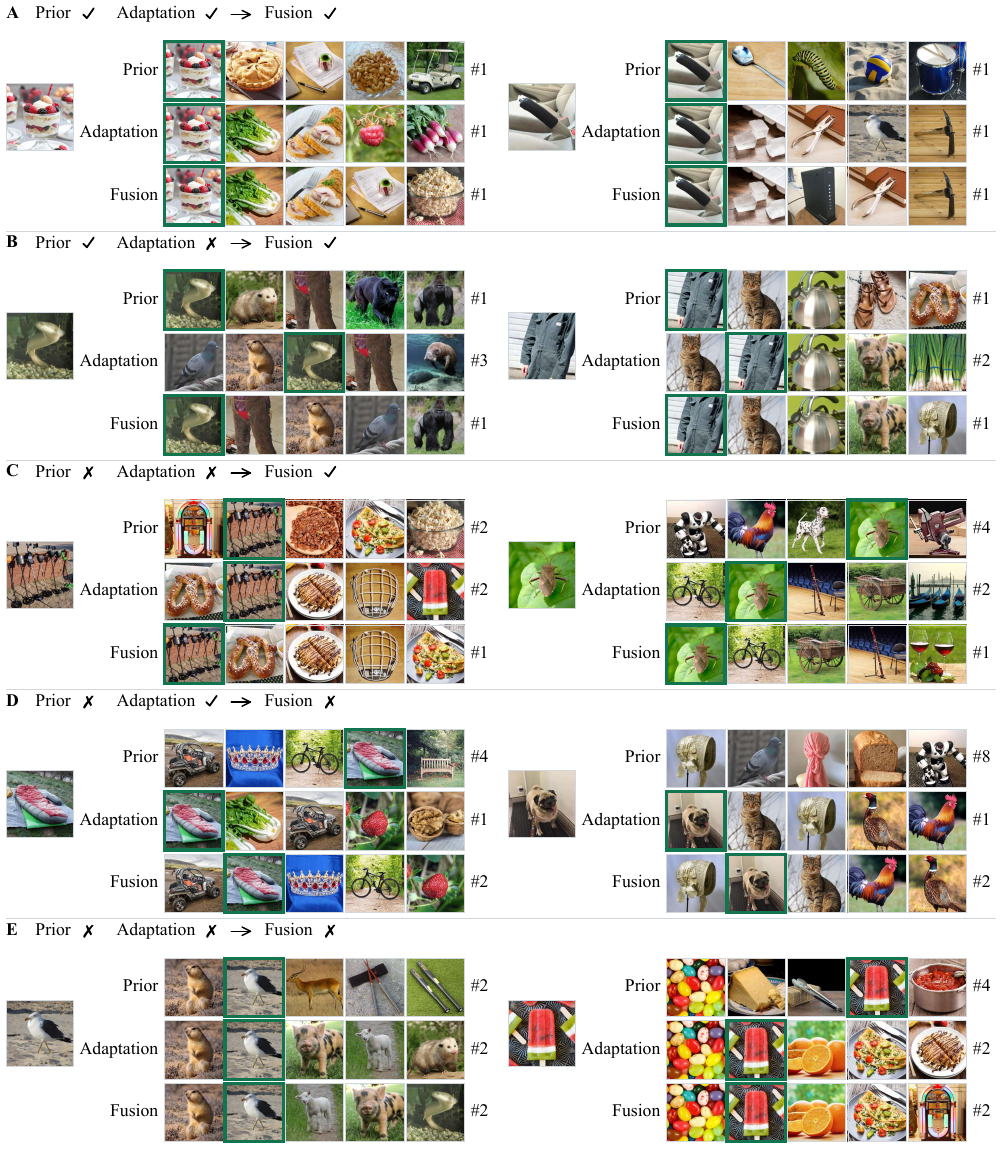}
    \caption{\textbf{Within-subject retrieval cases grouped by top-1 correctness.}
    In each group label, \ding{51} and \ding{55} mark whether the prior branch, the adaptation branch, and fusion rank the correct image first.
    Each row shows the top-5 retrievals of one model, and green boxes mark the correct image.
    \#k gives the rank of the correct image among all 200 candidates.}
    \label{fig:supp-fusion-cases}
    \end{minipage}
\end{figure*}

Fig.~\ref{fig:supp-fusion-cases} shows more within-subject retrieval cases, grouped by the top-1 correctness of the two branches and fusion.
Consistent with Fig.~8 of the main paper, fusion keeps the correct answer of the prior branch (group B) and corrects queries that both branches miss (group C), while groups D and E show some failure cases.

\DeclareRobustCommand{\suppFusionCorrect}{\ding{51}}
\DeclareRobustCommand{\suppFusionWrong}{\ding{55}}

\begin{table}[t]
    \retargettablefont
    \caption{\textbf{Within-subject top-1 results of fusion, grouped by the correctness of the two branches.}
    \suppFusionCorrect{} and \suppFusionWrong{} mark whether each branch alone retrieves the correct image at top-1.
    Fused correct counts the evaluations in each group that the fused model answers correctly.}
    \label{tab:supp-error-transitions}
    \centering
    \normalsize
    \setlength{\tabcolsep}{3pt}
    \renewcommand{\arraystretch}{1.15}
    \begin{tabular*}{\columnwidth}{@{\extracolsep{\fill}}cccc@{}}
        \toprule
        Prior & Adaptation & Count & Fused correct \\
        \midrule
        \suppFusionCorrect & \suppFusionCorrect & 6,974 & 6,974 \\
        \suppFusionWrong & \suppFusionCorrect & 1,932 & 1,806 \\
        \suppFusionCorrect & \suppFusionWrong & 666 & 554 \\
        \suppFusionWrong & \suppFusionWrong & 428 & 143 \\
        \midrule
        Total & & 10,000 & 9,477 \\
        \bottomrule
    \end{tabular*}
\end{table}

\subsection{Encoding Analysis Details}
\label{sec:supp-encoding-analysis}
Fig.~5 of the main paper uses encoding models to test how well each visual feature predicts the EEG response to the same image.
For each subject, electrode, and time point, we fit a ridge regression from one of five 768-d features to the EEG amplitude.
The features are Gabor, HOG, color, and the L06 and L11 layers of HA-SynCLR.
The analysis uses EEG without whitening and covers all 63 electrodes from $-100$ to 496 ms at 4 ms intervals.
The ridge penalty is selected on a validation split grouped by training concept, and the model is then refit on all 16,540 training images.
We evaluate each model by the Pearson correlation between predicted and measured EEG over the 200 test images.
In Fig.~5a, dots on each curve mark time points where the group-mean correlation is significantly above zero (one-sided, $p\leq0.05$, max-T corrected across the five features and all post-stimulus time points).

Tab.~\ref{tab:supp-encoding-analysis} summarizes the predictions at Oz and POz.
We focus on 60--100 ms because this window overlaps the latency of early visual evoked responses to patterned stimuli~\cite{dirusso2002vep}.
At these two occipital electrodes, Gabor predicts the EEG in this window about as well as HA-SynCLR L06 and clearly better than HOG and color.
Over the full time course, the group-mean curve of Gabor peaks earlier than those of both deep layers, while L06 reaches the highest peak correlation.
Color peaks earliest, but its correlation stays low throughout.

\begin{table}[t]
    \retargettablefont
    \caption{\textbf{Encoding performance at Oz and POz.}
    60–100 ms \textit{r} is the mean held-out correlation in this window, and ± is the standard deviation across subjects.
    Peak time and peak \textit{r} give the maximum of the group-mean curve plotted in Fig.~5a of the main paper.}
    \label{tab:supp-encoding-analysis}
    \centering
    \normalsize
    \setlength{\tabcolsep}{1pt}
    \renewcommand{\arraystretch}{1.15}
    \begin{tabular*}{\columnwidth}{@{\extracolsep{\fill}}lccc@{}}
        \toprule
        Feature & 60–100 ms \textit{r} & Peak time (ms) & Peak \textit{r} \\
        \midrule
        Gabor & 0.493 ± 0.072 & 104 & 0.699 \\
        HOG & 0.269 ± 0.044 & 116 & 0.501 \\
        Color & 0.246 ± 0.066 & 72 & 0.287 \\
        HA-SynCLR L06 & 0.483 ± 0.122 & 116 & 0.753 \\
        HA-SynCLR L11 & 0.432 ± 0.126 & 116 & 0.703 \\
        \bottomrule
    \end{tabular*}
\end{table}

\subsection{Pixel, DoG, and Gabor Targets}
\label{sec:supp-pixel-dog-gabor}
Fig.~2 of the main paper shows that Gabor is the strongest single descriptor.
To see where this decodability comes from, we compare three 768-d targets that describe the image with increasing local structure.
Grayscale pixels keep the intensity at every location but involve no filtering.
DoG responses measure the contrast between a local center and its surround, regardless of orientation.
Gabor responses also encode the orientation and spatial frequency of local contrast.
For the pixel target, we apply PCA fit on the training images.
The DoG responses are pooled over the same $4\times4$ regions as the Gabor descriptor, and the Gabor target is the single-descriptor target in Fig.~2.
The EEG encoder is trained on each target in the same way as the prior branch.

As shown in Tab.~\ref{tab:supp-pixel-dog-gabor}, grayscale pixels are barely decodable.
DoG raises top-1 accuracy by about 40 points.
Gabor brings a further gain over DoG.
Both gains are significant across the ten subjects, with $p=0.0020$ and $p=0.0039$ in a two-sided exact sign-flip test with max-T correction.
This indicates that the decodability of Gabor targets comes mainly from local filtering rather than from raw pixel values.
Adding orientation and spatial-frequency information further improves decoding.

\begin{table}[t]
    \retargettablefont
    \caption{\textbf{Within-subject retrieval accuracy (\%) with three 768-d visual targets.}
    All targets use the same EEG encoder and training procedure.}
    \label{tab:supp-pixel-dog-gabor}
    \centering
    \normalsize
    \setlength{\tabcolsep}{3pt}
    \renewcommand{\arraystretch}{1.15}
    \begin{tabular*}{\columnwidth}{@{\extracolsep{\fill}}lcc@{}}
        \toprule
        Target & Top-1 & Top-5 \\
        \midrule
        Grayscale pixels & 6.52 ± 0.55 & 19.57 ± 0.68 \\
        DoG & 46.65 ± 0.48 & 77.51 ± 0.52 \\
        \textbf{Gabor} & \textbf{52.42 ± 0.70} & \textbf{81.87 ± 0.43} \\
        \bottomrule
    \end{tabular*}
\end{table}

\section{Additional Results}
\label{sec:supp-additional-results}

\subsection{Per-subject Comparison}
\label{sec:supp-per-subject}
Tab.~\ref{tab:supp-per-subject} reports the per-subject results of the methods in Tab.~1 of the main paper.
In the within-subject setting, ReTarget achieves the highest top-1 and top-5 accuracy on every subject.
Its top-1 accuracy exceeds that of the second-best method on each subject by 5.30 to 14.70 points.
In the LOSO setting, ReTarget achieves the highest accuracy on nine of ten held-out subjects for both metrics.
Its advantage is clearest on Subjects 1 and 10, where top-1 accuracy exceeds 43\% and is about 15 and 9 points above the second-best method.
These results show that ReTarget is effective for each subject and maintains its advantage on subjects unseen during training.

\begin{table*}[t]
    \retargettablefont
    \caption{\textbf{Per-subject top-1 and top-5 accuracy (\%) of 200-way retrieval on THINGS-EEG.}
    (a) Within-subject setting.
    (b) LOSO setting, where each column denotes the held-out test subject.
    Bold marks the best result for each subject.}
    \label{tab:supp-per-subject}
    \centering
    \normalsize
    \setlength{\tabcolsep}{2pt}
    \renewcommand{\arraystretch}{1.1}
    \begin{tabular*}{\textwidth}{@{\extracolsep{\fill}}lc*{11}{c}@{}}
        \toprule
        \multicolumn{13}{c}{\refstepcounter{subtable}\label{tab:supp-per-subject-within}\textbf{(a) Within-subject: train and test on one subject}} \\
        \midrule
        Method & Metric & \makebox[25pt][c]{Sub1} & \makebox[25pt][c]{Sub2} & \makebox[25pt][c]{Sub3} & \makebox[25pt][c]{Sub4} & \makebox[25pt][c]{Sub5} & \makebox[25pt][c]{Sub6} & \makebox[25pt][c]{Sub7} & \makebox[25pt][c]{Sub8} & \makebox[25pt][c]{Sub9} & \makebox[25pt][c]{Sub10} & \makebox[25pt][c]{Avg.} \\
        \midrule
        NICE & Top-1 & 13.20 & 13.50 & 14.50 & 20.60 & 10.10 & 16.50 & 17.00 & 22.90 & 15.40 & 17.40 & 16.10 \\
        & Top-5 & 39.50 & 40.30 & 42.70 & 52.70 & 31.50 & 44.00 & 42.10 & 56.10 & 41.60 & 45.80 & 43.60 \\
        ATM & Top-1 & 25.60 & 22.00 & 25.00 & 31.40 & 12.90 & 21.30 & 30.50 & 38.80 & 34.40 & 29.10 & 27.10 \\
        & Top-5 & 60.40 & 54.50 & 62.40 & 60.90 & 43.00 & 51.10 & 61.50 & 72.00 & 51.50 & 63.50 & 58.10 \\
        Neural-MCRL & Top-1 & 27.50 & 28.50 & 37.00 & 35.00 & 22.50 & 31.50 & 31.50 & 42.00 & 30.50 & 37.50 & 32.40 \\
        & Top-5 & 64.00 & 61.50 & 69.00 & 66.00 & 51.50 & 61.00 & 62.50 & 74.50 & 59.50 & 71.00 & 64.10 \\
        UBP & Top-1 & 41.20 & 51.20 & 51.20 & 51.10 & 42.20 & 57.50 & 49.00 & 58.60 & 45.10 & 61.50 & 50.90 \\
        & Top-5 & 70.50 & 80.90 & 82.00 & 76.90 & 72.80 & 83.50 & 79.90 & 85.80 & 76.20 & 88.20 & 79.70 \\
        NeuroBridge & Top-1 & 50.00 & 63.20 & 61.60 & 61.40 & 54.80 & 69.70 & 62.70 & 71.20 & 64.00 & 73.60 & 63.20 \\
        & Top-5 & 77.60 & 90.60 & 91.10 & 90.00 & 85.00 & 92.90 & 88.80 & 95.10 & 91.00 & 97.10 & 89.90 \\
        Shallow Alignment & Top-1 & 75.00 & 87.50 & 83.20 & 79.50 & 74.60 & 89.90 & 78.50 & 86.90 & 81.30 & 89.30 & 82.60 \\
        & Top-5 & 94.30 & 98.90 & 98.20 & 96.10 & 96.40 & 99.30 & 97.30 & 99.40 & 97.80 & 99.10 & 97.70 \\
        HCF & Top-1 & 81.90 & 86.00 & 84.30 & 82.20 & 74.50 & 90.40 & 82.80 & 90.90 & 79.70 & 93.10 & 84.60 \\
        & Top-5 & 98.30 & 98.90 & 97.70 & 98.80 & 94.10 & 98.90 & 97.20 & 99.40 & 98.40 & 99.80 & 98.20 \\
        ReTarget & Top-1 & \textbf{93.80} & \textbf{96.00} & \textbf{96.20} & \textbf{89.20} & \textbf{87.60} & \textbf{97.50} & \textbf{96.10} & \textbf{96.90} & \textbf{96.00} & \textbf{98.40} & \textbf{94.77} \\
        & Top-5 & \textbf{99.10} & \textbf{99.90} & \textbf{100.00} & \textbf{99.40} & \textbf{98.90} & \textbf{100.00} & \textbf{100.00} & \textbf{100.00} & \textbf{100.00} & \textbf{100.00} & \textbf{99.73} \\
        \midrule
        \multicolumn{13}{c}{\refstepcounter{subtable}\label{tab:supp-per-subject-loso}\textbf{(b) LOSO: leave one subject out for test}} \\
        \midrule
        Method & Metric & \makebox[25pt][c]{Sub1} & \makebox[25pt][c]{Sub2} & \makebox[25pt][c]{Sub3} & \makebox[25pt][c]{Sub4} & \makebox[25pt][c]{Sub5} & \makebox[25pt][c]{Sub6} & \makebox[25pt][c]{Sub7} & \makebox[25pt][c]{Sub8} & \makebox[25pt][c]{Sub9} & \makebox[25pt][c]{Sub10} & \makebox[25pt][c]{Avg.} \\
        \midrule
        NICE & Top-1 & 7.60 & 5.90 & 6.00 & 6.30 & 4.40 & 5.60 & 5.60 & 6.30 & 5.70 & 8.40 & 6.20 \\
        & Top-5 & 22.80 & 20.50 & 22.30 & 20.70 & 18.30 & 22.20 & 19.70 & 22.00 & 17.60 & 28.30 & 21.40 \\
        ATM & Top-1 & 10.50 & 7.10 & 11.90 & 14.70 & 7.00 & 11.10 & 16.10 & 15.00 & 4.90 & 20.50 & 11.90 \\
        & Top-5 & 26.80 & 24.80 & 33.80 & 39.40 & 23.90 & 35.80 & 43.50 & 40.30 & 22.70 & 46.50 & 33.80 \\
        Neural-MCRL & Top-1 & 13.00 & 12.00 & 14.50 & 12.50 & 11.50 & 13.50 & 14.00 & 18.50 & 13.50 & 17.00 & 14.00 \\
        & Top-5 & 31.50 & 30.50 & 35.50 & 35.50 & 29.00 & 35.50 & 36.00 & 38.50 & 32.50 & 39.00 & 34.30 \\
        UBP & Top-1 & 11.50 & 15.50 & 9.80 & 13.00 & 8.80 & 11.70 & 10.20 & 12.20 & 15.50 & 16.00 & 12.40 \\
        & Top-5 & 29.70 & 40.00 & 27.00 & 32.30 & 33.80 & 31.00 & 23.80 & 32.20 & 40.50 & 43.50 & 33.40 \\
        NeuroBridge & Top-1 & 23.20 & 21.20 & 13.20 & 17.00 & 14.50 & 25.00 & 15.30 & 20.10 & 13.70 & 27.20 & 19.00 \\
        & Top-5 & 52.40 & 49.30 & 36.50 & 45.30 & 37.70 & 55.00 & 45.10 & 44.90 & 36.50 & 56.30 & 45.90 \\
        Shallow Alignment & Top-1 & 23.50 & 30.60 & 10.00 & 19.50 & 18.10 & 22.70 & 18.60 & 17.30 & \textbf{23.00} & 34.40 & 21.80 \\
        & Top-5 & 53.20 & \textbf{60.40} & 28.10 & 48.30 & 45.20 & 49.80 & 46.00 & 46.10 & 54.80 & 62.00 & 49.40 \\
        HCF & Top-1 & 28.50 & 22.50 & 20.00 & 20.50 & 18.50 & 29.50 & 20.00 & 20.00 & 21.00 & 34.00 & 23.40 \\
        & Top-5 & 61.00 & 56.50 & 45.50 & 54.00 & 46.50 & 60.50 & 53.50 & 53.50 & 48.50 & 69.00 & 54.90 \\
        ReTarget & Top-1 & \textbf{43.30} & \textbf{31.00} & \textbf{21.60} & \textbf{23.40} & \textbf{22.40} & \textbf{33.90} & \textbf{29.90} & \textbf{27.90} & 22.80 & \textbf{43.80} & \textbf{30.00} \\
        & Top-5 & \textbf{75.40} & 59.00 & \textbf{48.80} & \textbf{54.60} & \textbf{51.60} & \textbf{68.40} & \textbf{59.70} & \textbf{55.20} & \textbf{55.10} & \textbf{77.00} & \textbf{60.48} \\
        \bottomrule
    \end{tabular*}
\end{table*}

\subsection{Generalization to THINGS-MEG}
\label{sec:supp-meg}
To test whether ReTarget generalizes beyond EEG to other neural recordings, we apply it to THINGS-MEG~\cite{hebart2023thingsdata}, which records MEG responses to natural object images from four subjects.
We use all 271 channels from 0 to 500 ms at 200 Hz.
Each subject has 19,848 training images and 200 test images.
As for EEG, MEG responses are averaged over the repeated presentations of each image.
The LOSO setting trains on three subjects and tests on the remaining one.
The neural encoders, the visual targets of both branches, and all hyperparameters are the same as for EEG and are not tuned on MEG.

As shown in Tab.~\ref{tab:supp-meg}, ReTarget can be applied to MEG without retuning and shows the same pattern as on EEG.
Within subjects, the adaptation branch outperforms the prior branch, and fusing the two branches raises top-1 accuracy by a further 7.62 points.
This indicates that the explicit prior and the adapted deep target remain complementary on MEG.
In the LOSO setting, ReTarget still achieves the highest accuracy among the three models, but the fusion gain is limited.
As on EEG, fusion helps less across subjects than within subjects.

\begin{nolinenumbers}
\par\medskip\noindent
\begin{minipage}{\columnwidth}
    \captionsetup{hypcap=false}
    \retargettablefont
    \captionof{table}{\textbf{Top-1 and top-5 accuracy (\%) of 200-way retrieval on THINGS-MEG in the within-subject and LOSO settings.}}
    \label{tab:supp-meg}
    \centering
    \normalsize
    \setlength{\tabcolsep}{1pt}
    \renewcommand{\arraystretch}{1.15}
    \newcommand{\suppMEGStdDev}[1]{{\fontsize{8}{10}\selectfont ±\,#1}}
    \begin{tabular*}{\columnwidth}{@{\extracolsep{\fill}}lcccc@{}}
        \toprule
        \multirow{2}{*}{Method} & \multicolumn{2}{c}{Within-subject} & \multicolumn{2}{c}{LOSO} \\
        \cmidrule(lr){2-3}\cmidrule(l){4-5}
        & Top-1 & Top-5 & Top-1 & Top-5 \\
        \midrule
        Prior & 23.38\suppMEGStdDev{0.49} & 46.17\suppMEGStdDev{1.88} & 2.55\suppMEGStdDev{0.26} & 10.88\suppMEGStdDev{0.41} \\
        Adaptation & 36.48\suppMEGStdDev{0.82} & 63.67\suppMEGStdDev{0.76} & 4.33\suppMEGStdDev{0.26} & 16.03\suppMEGStdDev{0.91} \\
        \textbf{ReTarget} & \textbf{44.10\suppMEGStdDev{0.55}} & \textbf{69.95\suppMEGStdDev{0.86}} & \textbf{4.48\suppMEGStdDev{0.55}} & \textbf{17.05\suppMEGStdDev{0.54}} \\
        \bottomrule
    \end{tabular*}
\end{minipage}
\par\end{nolinenumbers}

\clearpage
\end{document}